%% file: arxiv.tex
\documentclass[letterpaper]{article}

\usepackage[preprint]{aaai2027}
\usepackage[hyphens]{url}
\usepackage{graphicx}
\usepackage{natbib}
\usepackage{caption}
\usepackage{booktabs}
\usepackage{multirow}
\usepackage{amsmath}
\usepackage{amssymb}
\usepackage[table]{xcolor}
\usepackage{enumitem}
\usepackage{xspace}
\usepackage{tabularx}
\usepackage[most]{tcolorbox}
\usepackage{listings}
\usepackage{microtype}

\newcolumntype{Y}{>{\centering\arraybackslash}X}
\newcommand{\imp}[1]{\textcolor{green!60!black}{\scriptsize{\,(+#1)}}}


\newcommand{\method}{OpenVisTool\xspace}
\newcommand{\dataset}{OpenVisTool-42K\xspace}
\newcommand{\benchname}{OpenVisTool-Bench\xspace}

\newcounter{internfn}
\newcommand{\internship}{%
  \ifnum\value{internfn}=0%
    \footnote{Work done during internships at SenseTime Research.}%
    \setcounter{internfn}{\value{footnote}}%
  \else%
    \footnotemark[\value{internfn}]%
  \fi%
}

\title{OpenVisTool: An Open Recipe for Synthesizing \\
Instructive Visual Tool-Use Trajectories%
}

\author{%
Changhao Xiang\textsuperscript{\rm 1,2}\equalcontrib\internship,
Shilin Zhang\textsuperscript{\rm 1,2}\equalcontrib\internship,
Zheng Ma\textsuperscript{\rm 2}\thanks{Project leader.}
Kanzhi Cheng\textsuperscript{\rm 1},
Ruize Ma\textsuperscript{\rm 1},
Yi Feng\textsuperscript{\rm 1},\\
Jianbing Zhang\textsuperscript{\rm 1},
Zhi Wang\textsuperscript{\rm 1}\corresponding,
Zhen Wu\textsuperscript{\rm 1}\corresponding,
Xinyu Dai\textsuperscript{\rm 1},
Lewei Lu\textsuperscript{\rm 2}
}
\affiliations{%
\textsuperscript{\rm 1}Nanjing University
\quad
\textsuperscript{\rm 2}SenseTime Research\\
\texttt{\{xiangch,shilinzhang\}@smail.nju.edu.cn}
\quad
\texttt{\{zhiwang,wuz\}@nju.edu.cn}
}

\begin{document}

\maketitle

\input{sections/introduction}
\input{sections/related_work}
\input{sections/method}
\input{sections/experiments}
\input{sections/conclusion}

% aaai2027.sty already sets \bibliographystyle{aaai2027}.
\bibliography{custom}

\clearpage
\appendix
\section*{Supplementary Material}
\input{sections/appendix}

\end{document}

%% file: sections/introduction.tex
\begin{abstract}
Visual tool use has emerged as a fundamental capability for multimodal agents to actively acquire evidence beyond a fixed image encoding. 
The prevailing recipe learns this capability from teacher-generated trajectories filtered for answer correctness, implicitly assuming that every successful demonstration provides effective supervision. 
We argue this assumption is flawed: a strong teacher often reaches the correct answer without needing its tool calls, and imitating such trajectories teaches a student that tool calls accompany correct answers, not that tool observations ground them. 
We present \textbf{OpenVisTool}, an open framework for constructing \textit{instructive visual tool-use trajectories} that provide effective supervision for tool learning.
The key insight is that a trajectory should be retained only if its answer is correct (outcome validity) and its tool observations causally contribute to that answer (causal utility).
The framework operates in three stages: \textit{difficulty screening} to select queries that are not reliably answerable without tools, \textit{domain-specific trajectory synthesis} to elicit coherent tool-use trajectories, and \textit{supervision verification} to jointly test both conditions.
Rather than encouraging models to imitate tool calls, the resulting supervision teaches when and how visual evidence should be acquired.
Using this framework, we construct \textbf{\dataset}, a dataset spanning five visual reasoning domains, together with \textbf{\benchname}, a benchmark covering the same domains.
Across four backbones (4B–27B), fine-tuning on \dataset consistently improves visual tool-use performance and yields gains on two out-of-distribution benchmarks; the larger models approach leading closed-source systems.
The evidence suggests that effective visual tool use is learned from causally grounded supervision rather than tool-calling patterns.
Our code is available at \url{https://github.com/Changhao-Xiang/OpenVisTool}.

\end{abstract}

\begin{figure}[t]
    \centering

    \includegraphics[width=\columnwidth]{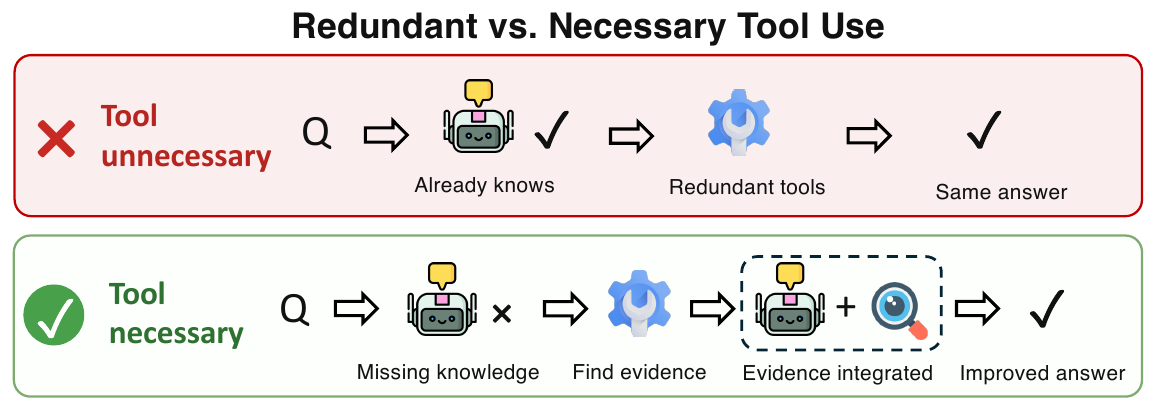}

    \vspace{2mm}
    % \hrule
    % \vspace{2mm}
    
    \includegraphics[width=0.9\columnwidth]{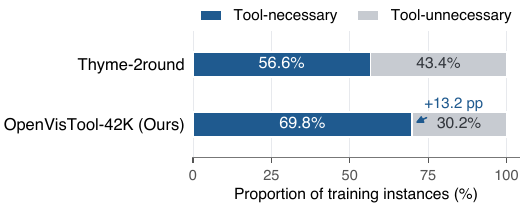}
    \caption{Correct tool-use trajectories are not always instructive. (Top) Illustration of tool-unnecessary and tool-necessary trajectories. (Bottom) Tool-necessity distributions for Thyme-2round and our \dataset, computed over 1,000 instances by re-querying the corresponding teacher model without tool access. Trajectories with correct no-tool answers are labeled \textit{tool-unnecessary}.}
    \label{fig:tool_necessity}
\end{figure}

\section{Introduction}
\label{sec:introduction}

Large language models become more capable when equipped with external tools such as code interpreters \citep{xue2025simpletir,feng2025retool} and search engines \citep{jin2025searchr1,Search-o1,Li2025WebThinker}. 
For multimodal models, the bottleneck is visual: encoding an image into a fixed set of tokens inevitably discards the fine-grained evidence that many questions hinge on~\citep{su2025thinkingsurvey}.
Visual tools recover such evidence by feeding new observations into reasoning--cropping to reveal small text \citep{chartagent}, zooming in to resolve spatial details \citep{zheng2026deepeyes}, or rendering generated code against a reference image \citep{ui2coden2026}--a paradigm known as \textit{thinking with images}~\citep{openai2025o3}.
To instill this capability, recent systems couple cold-start supervised fine-tuning (SFT) on teacher-generated trajectories with reinforcement learning (RL) that refines the policy \citep{lai2026minio3,hou2025codev,guo2025codevision}.
Since RL alone rarely bootstraps stable tool invocation \citep{hong2026deepeyesv2,zhang2026thyme}, the quality of the cold-start data largely determines a model's ultimate tool-use competence.
However, not all trajectories contribute equally, and \textit{what makes one genuinely valuable for learning} remains an open question.

Existing synthesis pipelines typically begin with outcome-based rejection sampling: a trace is retained if it is executable, well-formed, and yields a correct answer \citep{lai2026minio3,hong2026deepeyesv2}.
This criterion, however, conflates outcome validity with the utility of the tool calls behind it.
Some pipelines further employ model or human judges to verify that tool outputs are consistent with the reasoning and plausibly support the answer \citep{zhang2026thyme,zhang2025skyworkr1v4}. Yet such judgments still do not establish counterfactual utility.
A strong teacher can often reach the right answer without tools (via parametric knowledge, reasoning shortcuts, or coarse image cues), yet still issues tool calls simply because the synthesis prompt instructs it to~\citep{hou2025codev,zhao2026focus}.
Such trajectories pass both outcome- and judge-based filtering while providing \textit{spurious supervision}: the visual observations appear relevant but make no causal contribution to the answer. A student trained on such data learns correlation rather than causation: it learns that tool calls \textit{accompany} correct answers, not that tool observations \textit{ground} them.
Answering the open question above thus requires shifting the object of evaluation from whether a trajectory ends correctly to whether its tool calls do real work: \textit{would the answer still hold if the visual observations were taken away?}

Our answer is a two-part criterion: a trajectory constitutes beneficial supervision for visual tool-use learning only when it satisfies two conditions jointly. 
The first is \textit{outcome validity}: the trajectory must reach a correct final answer, so that the model is not trained on erroneous reasoning chains. 
The second, which existing pipelines overlook, is \textit{causal utility}: the trajectory's tool trace must improve a fixed probe model's answer reliability relative to a no-tool baseline, with the original image and query held constant.
We estimate this utility by comparing the probe model's success rate when conditioned on the recorded tool calls and observations against its success rate without them.
A trajectory meeting both conditions teaches the model not merely that tools \textit{can} be called, but \textit{why} a call is worth making: it supplies visual evidence that the model's intrinsic encoding cannot.
This criterion is not merely conceptual but directly testable. We build our data synthesis pipeline around it.

Guided by this criterion, we design \textbf{OpenVisTool}, a three-stage trajectory synthesis pipeline that puts causal supervision selection into practice across five visual reasoning domains (chart, table, GUI grounding, visual search, and web-to-HTML) under a shared visual toolset. 
First, we apply \textit{Difficulty Screening} to retain only questions that a capable model fails to answer reliably without tool access, so that the surviving problems genuinely demand external visual evidence.
Second, during \textit{Domain-Specific Trajectory Synthesis}, we use structured invocation strategies tailored to each domain to guide the teacher toward coherent traces rather than arbitrary tool calls.
Finally, we perform \textit{Supervision Verification} to ensure that each trajectory satisfies both conditions above: it is retained only if its final answer is correct (outcome validity) and its tool traces, when provided to a base model, improve its answer (causal utility).
Together, the three stages impose progressively stricter requirements: a question must demand visual tools, a trace must invoke them coherently, and an answer must causally depend on what the tools return.
In summary, our contributions are threefold:
% \begin{itemize}
%     \item We propose \textbf{\method}, a data synthesis framework that selects visual tool-use trajectories by the causal contribution of tool observations to task success, and use it to construct \textbf{\dataset}, an open-source large-scale dataset spanning five representative domains.
%     \item We introduce \textbf{\benchname}, a comprehensive benchmark for visual tool use covering the same five domains, which supports standardized and reproducible evaluation of both open-source and proprietary models.
%     \item We fine-tune visual tool agents on four backbones (4B–27B) using \dataset. The resulting models are competitive with leading closed-source models, generalize to out-of-distribution tasks, and outperform models trained with alternative data-filtering strategies.
% \end{itemize}

\begin{itemize}
    \item We introduce \textbf{\method}, a three-stage framework for constructing instructive visual tool-use trajectories by jointly enforcing outcome validity and causal utility.
    \item We construct \textbf{\dataset}, an open-source large-scale dataset spanning five representative visual reasoning domains under a shared visual toolset.
    \item We fine-tune visual tool agents on four backbones (4B–27B) using \dataset. The resulting models are competitive with leading closed-source models, generalize to out-of-distribution tasks, and outperform models trained with alternative data-filtering strategies.
\end{itemize}

%% file: sections/related_work.tex
\section{Related Work}
\label{sec:related_work}

\paragraph{Visual Tool-Use Learning}
Recent work on training visual tool-use agents follows two paradigms. One directly optimizes tool-calling policies via reinforcement learning \citep{zheng2026deepeyes, wu2026vtoolr}, but RL alone often fails to bootstrap stable tool invocation without a well-initialized policy. The dominant recipe therefore adopts a two-stage approach: cold-start supervised fine-tuning (SFT) on teacher-generated trajectories to establish basic tool-use competence, followed by RL for further refinement \citep{su2025openthinkimg, su2026pixelreasoner, hong2026deepeyesv2, lai2026minio3, guo2025codevision, hou2025codev}. Under this paradigm, SFT quality directly determines whether RL can converge to capable policies \citep{hong2026deepeyesv2, zhang2026thyme}. Yet existing work largely treats the SFT corpus as a commodity—any answer-correct teacher trajectory is assumed to provide useful supervision. What constitutes genuinely beneficial supervision for visual tool-use learning has not been systematically studied.

\paragraph{Supervision Quality for Reasoning}
In text-based reasoning, the quality of training data has received sustained attention. Rejection sampling—retaining only solutions that reach the correct final answer—is a standard recipe for curating reasoning corpora \citep{zelikman2022star, yuan2023scaling}. Subsequent studies demonstrate that not all correct traces are equally instructive: process-level verification reveals that individual reasoning steps vary in correctness and informativeness \citep{lightman2024let, wang2024math}, and difficulty-aware or diversity-aware selection further improves learning efficiency over naive outcome filtering \citep{yu2024metamath}. These findings establish a clear lesson: supervision quality matters as much as quantity. However, this lesson has yet to be transferred to the visual tool-use setting, where the notion of ``quality'' is further complicated by tool observations—a returned modality whose actual contribution to reasoning is neither guaranteed nor straightforward to assess.

\begin{figure*}
    \centering
    \includegraphics[width=\textwidth]{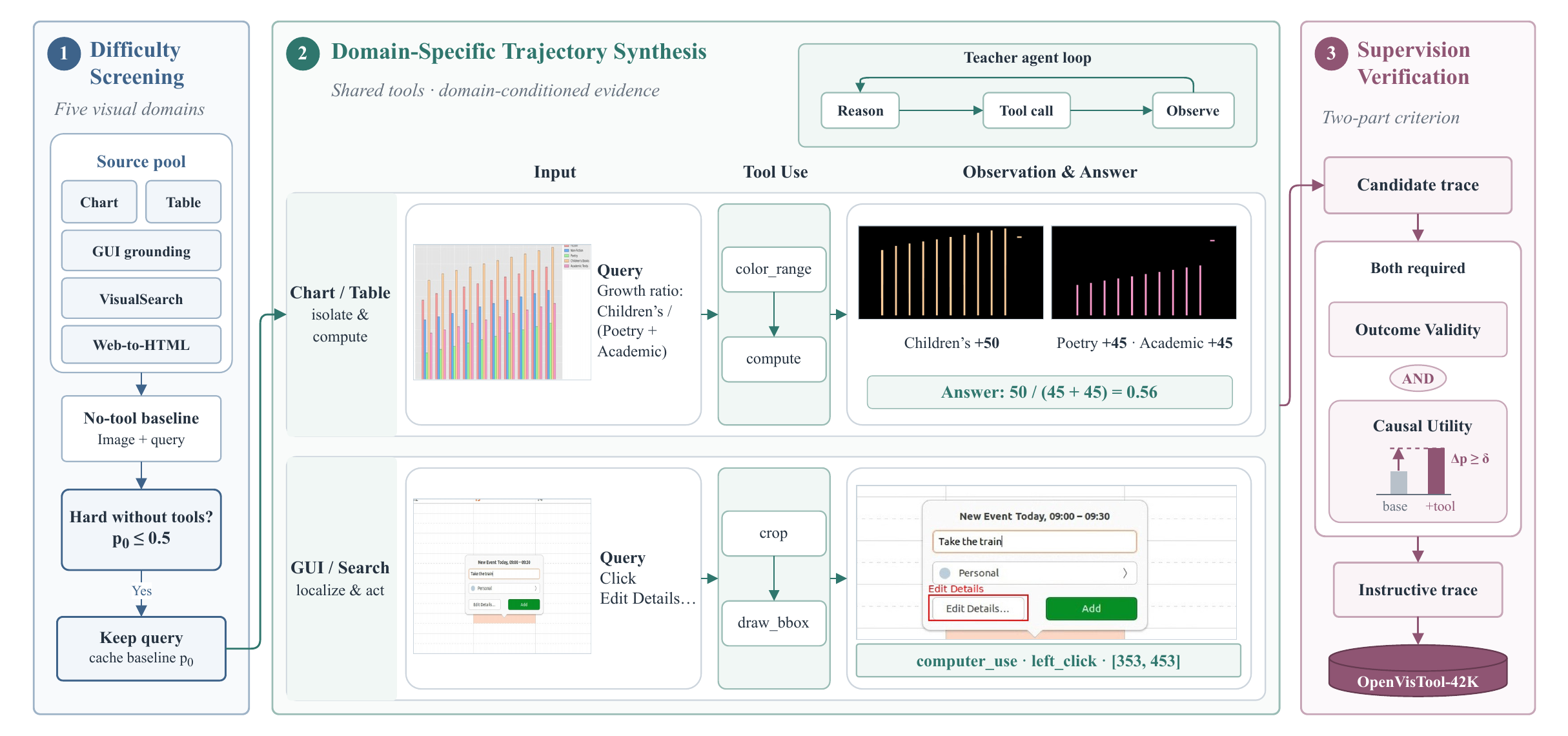}
    \caption{Overview of the three-stage \method data pipeline. \textbf{Difficulty Screening} identifies queries that the probe model cannot reliably solve from the original image alone. \textbf{Domain-Specific Trajectory Synthesis} uses domain-specific tool-use patterns to guide a stronger teacher in acquiring evidence that addresses each domain's visual bottleneck. \textbf{Supervision Verification} determines whether each trajectory is instructive: it must achieve a valid outcome, and its tool observations must causally improve the probe model's answer reliability. Only trajectories satisfying both criteria are retained.}
    \label{fig:pipeline}
\end{figure*}

\paragraph{Visual Evidence Acquisition}
The paradigm of ``thinking with images,'' introduced by OpenAI's o3 \citep{openai2025o3}, reframes visual reasoning as an active evidence-acquisition process: rather than relying on a single static encoding, models iteratively crop, zoom, or render visual inputs and incorporate the resulting observations into their reasoning \citep{su2025thinkingsurvey}. Under this lens, visual tools acquire evidence that the model's intrinsic visual encoding cannot provide. This perspective motivates our causal framing: a trajectory is instructive only when its tool observations causally contribute to reaching the correct answer. Trajectories in which tools are invoked but the answer could have been obtained without the returned observations teach \emph{correlation}—tool calls co-occur with success—rather than the deeper skill of knowing \emph{when and why} to seek external visual evidence.

%% file: sections/method.tex
\section{\method}
\label{sec:method}

\subsection{What Makes a Trajectory Instructive?}
\label{sec:instructive-trajectory}

% 这里做一些 problem formulation，并用公式化表达
Consider a visual reasoning problem with input image $I$, query $q$, and reference answer $y^\star$. A candidate tool-use trajectory is represented as
\begin{equation}
\tau=\left((r_t,c_t,o_t)_{t=1}^{T},\hat{y}_{\tau}\right),
\end{equation}
where $T$ is the number of turns, $r_t$, $c_t$, and $o_t$ denote the reasoning step, tool call, and observation at turn $t$, respectively; and $\hat{y}_{\tau}$ is the final answer. 

An instructive trajectory must satisfy two complementary requirements.
\textbf{Outcome Validity} requires the trajectory to terminate in a correct answer under the task-specific evaluator.
However, correctness alone does not make it instructive: a strong teacher may infer the answer from parametric knowledge or reasoning shortcuts while still invoking tools because the synthesis prompt encourages it.
Such a trajectory is successful in outcome but spurious as supervision: it teaches that tool calls co-occur with correct answers, without establishing that the tool interactions are useful.

\textbf{Causal Utility} therefore requires the recorded tool calls and observations to improve a fixed probe model's answer reliability relative to a no-tool condition, while the original image and query remain unchanged.
Sections~\ref{sec:data} and~\ref{sec:filtering} operationalize these requirements through repeated stochastic probe trials.
Figure~\ref{fig:pipeline} turns them into a three-stage pipeline. 
\textbf{Difficulty Screening} first selects queries that the probe model cannot reliably solve from the original image alone.
\textbf{Domain-Specific Trajectory Synthesis} then uses domain-specific tool-use patterns to guide a stronger teacher toward acquiring evidence suited to each domain's visual bottleneck.
Finally, \textbf{Supervision Verification} evaluates outcome validity and estimates causal utility by comparing the probe model under tool-trace-conditioned and no-tool conditions.
Only trajectories satisfying both criteria are retained to form OpenVisTool-42K.

\subsection{Difficulty Screening}
\label{sec:data}

We collect candidate instances from five domains: Chart, Table, GUI Grounding, Visual Search, and Web-to-HTML. 
Since many queries can already be solved from the original image, synthesizing tool-use trajectories for them may introduce redundant tool calls. 
We therefore screen for queries that a fixed probe model cannot reliably answer without tools.
% 这里面把 four times 换成 K 次，更加通用正式一些，实验中 K = 4
% 成功最多两次也换成了成功率不超过gamma
Let $x=(I,q,y^\star,d)$ denote a candidate instance, where $d$ is the task domain, and let $E_d(\hat{y},y^\star)\in\{0,1\}$ denote the corresponding domain-specific evaluator.
For each $x$, we run a fixed probe model $\pi_0$ for $K$ independent no-tool trials.
Let $Y_{\pi_0}^{(k)}(I,q)$ denote its answer in trial $k$.
We compute the empirical success rate
\begin{equation}
\bar{p}_0(x)=\frac{1}{K}\sum_{k=1}^{K}E_d\left(Y_{\pi_0}^{(k)}(I,q),y^\star\right).
\label{eq:no-tool-success-rate}
\end{equation}
For difficulty threshold $\gamma$, we retain an instance if $\bar{p}_0(x)\leq\gamma$.
We use Qwen3.5-9B \citep{qwen35blog} as $\pi_0$ and set $K\!=\!4, \gamma\!=\!0.5$, retaining queries answered correctly in at most two trials. 
Dataset sources and preprocessing details are provided in Appendix A.

\subsection{Domain-Specific Trajectory Synthesis}
\label{sec:synthesis}

For each instance retained by Difficulty Screening, we use Qwen3.5-Plus as a teacher model $\pi_{\text{teacher}}$ to synthesize a candidate trajectory with a shared visual toolset. 
Let $h_t$ denote the interaction history before turn $t$. 
The rollout proceeds as
\begin{equation}
\begin{aligned}
(r_t,c_t)
&\sim \pi_{\mathrm{teacher}}(\cdot\mid h_t,P_d), \\
o_t
&=\operatorname{Exec}(c_t), \\
h_{t+1}
&=h_t\oplus(r_t,c_t,o_t),
\end{aligned}
\label{eq:teacher-rollout}
\end{equation}
where $P_d$ is the tool-use pattern associated with domain $d$. 
The process continues until the teacher produces a final answer, and the complete sequence of reasoning steps, tool calls, and observations is preserved.

A generic instruction to ``use tools when helpful'' may elicit arbitrary or decorative calls. We instead define $P_d$ around the dominant visual bottleneck of each domain: series isolation, comparison, and computation for Chart; relevant-cell localization and row--column alignment for Table; progressive target localization and verification for GUI Grounding and Visual Search; and iterative render--compare--revise for Web-to-HTML. All domains use the same underlying tool interface. Consequently, the synthesized trajectories emphasize reusable evidence-acquisition behaviors rather than domain-specific APIs. The full toolset and domain-specific instructions are provided in Appendices B and C, respectively.

\subsection{Supervision Verification}
\label{sec:filtering}

We first discard malformed trajectories, including sessions with missing observations, failed tool executions, or invalid file references. 
For each remaining trajectory, we verify \textbf{outcome validity} using the domain-specific evaluator $E_d$.
Only trajectories satisfying $E_d(\hat{y}_\tau,y^\star)=1$ proceed to the causal-utility test.
Depending on the domain, $E_d$ uses answer matching, point-in-box evaluation, or rendered-page comparison with a VLM judge.

We then evaluate \textbf{causal utility}. 
For each outcome-valid trajectory, let $Z_\tau=\bigl((c_t,o_t)\bigr)_{t=1}^{T}$ denote the sequence of tool calls and corresponding observations supplied, together with the original image and query, to the same probe model $\pi_0$ used in Difficulty Screening.
The teacher's reasoning and final answer are excluded to prevent solution leakage. 
Let $Y_{\pi_0}^{(k)}(I,q;Z_\tau)$ denote the probe's answer in tool-trace-conditioned trial $k$.
Using the same $K=4$ independent trials as in Difficulty Screening, we compute
\begin{equation}
\bar{p}_{\mathrm{tool}}(x,\tau)=\frac{1}{K}\sum_{k=1}^{K}E_d\left(Y_{\pi_0}^{(k)}(I,q;Z_\tau),y^\star
\right).
\label{eq:tool-trace-success-rate}
\end{equation}
For utility threshold $\delta$, a trajectory passes Supervision Verification when both
\begin{equation}
\begin{aligned}
E_d(\hat{y}_\tau,y^\star) &= 1, \\
\bar{p}_{\mathrm{tool}}(x,\tau)-\bar{p}_0(x) &\geq \delta
\end{aligned}
\label{eq:supervision-verification}
\end{equation}
hold.
We set $\delta=0.25$. 
Reusing the same probe model and evaluator makes $\bar{p}_{\mathrm{tool}}(x,\tau)$ directly comparable with $\bar{p}_0(x)$.
Together, the conditions in Eq.~\ref{eq:supervision-verification} retain only outcome-correct trajectories whose visual evidence measurably improves answer reliability.
Each verified trajectory forms a supervised fine-tuning sample that preserves the teacher's reasoning, tool calls, observations, and final answer, providing process-level supervision for when and how to use visual evidence. Domain-wise statistics for OpenVisTool-42K and qualitative examples are provided in Appendices A and D, respectively.
% Qualitative examples of the resulting trajectories are provided in Appendix D.

%% file: sections/experiments.tex
\input{tables/tab_main}

\section{Experiments}
\label{sec:experiments}

We design experiments to validate our central claim—that effective visual tool use is learned from instructive trajectories rather than merely successful ones—and to answer the following questions:
\begin{itemize}
    \item \textbf{Effectiveness:} Does fine-tuning on OpenVisTool-42K teach models to effectively leverage visual tools, and can it lift open-source models to closed-source performance levels? (\S\ref{sec:main_results})
    
    \item \textbf{Generalization:} Does the learned tool-use capability transfer to out-of-distribution visual tasks unseen during training, rather than overfitting to domain-specific tool-calling patterns? (\S\ref{sec:ood})
    
    \item \textbf{Supervision verification:} Are outcome validity and causal utility both necessary for constructing effective visual tool-use supervision? (\S\ref{sec:ablation_gain})
    
    \item \textbf{Cross-domain synergy:} Does visual tool use emerge as a transferable meta-skill across domains, and does multi-domain training resolve strategy conflicts that single-domain training introduces? (\S\ref{sec:cross_domain})
\end{itemize}

\input{sections/bench}

\subsection{Experimental Setup}
\label{sec:setup}

\paragraph{Benchmarks.}
% We primarily evaluate on \benchname, which covers diverse visual tool-use scenarios across multiple domains. 
We primarily evaluate on \benchname. To assess generalization, we additionally test on Agentic-MME and VTC-Bench. We also report results on the full Chart and Table source benchmarks in Appendix F to confirm that gains are not artifacts of sample selection.

\paragraph{Baselines.}
We compare three types of models. (i)~General-purpose frontier models: GPT-5.5 and Kimi K2.6~\citep{kimik26blog}, both evaluated without tools and with our shared toolset. (ii)~Smaller-scale general-purpose models: Qwen2.5-VL-7B~\citep{Qwen2.5-VL}, Qwen3-VL-8B-Instruct~\citep{Qwen3-VL} and Qwen3.5~\citep{qwen35blog} models are evaluated both without tools and with our shared toolset before fine-tuning. (iii)~Open-source visual tool-use agents: Thyme~\citep{zhang2026thyme} and DeepEyes V2~\citep{hong2026deepeyesv2}, evaluated with the code-based tool environments used in their original work rather than our toolset.

\paragraph{Training.}
We fine-tune Qwen3.5-4B/9B/27B and Qwen3-VL-8B-Instruct on \dataset for 3 epochs using SWIFT~\citep{zhao2025swift}. More implementation details are provided in Appendix G.

\input{tables/tab_ood}

\begin{figure}[t]
    \centering
    \includegraphics[width=0.9\columnwidth]{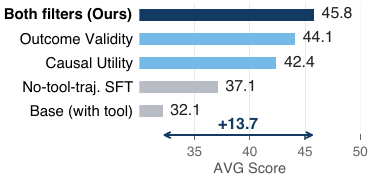}
    % \caption{Comparison of training variants.}
    \caption{Ablation of the criteria for instructive tool-use supervision on Qwen3.5-9B under a fixed training budget. Requiring both outcome validity and causal utility achieves the best performance, outperforming either criterion alone and the no-tool-trajectory SFT baseline.}
    \label{fig:ablation}
\end{figure}

\input{tables/tab_trajectory_replay}

\subsection{Main Results}
\label{sec:main_results}
Table~\ref{tab:main} summarizes performance across all models. We highlight several key findings below.

\paragraph{Consistent gains across all base models and domains.}
Our method yields substantial improvements over the no-tool baselines across all domains and model scales. Across models ranging from 4B to 27B, fine-tuning with \dataset{} brings an average gain of 10.7 points on \benchname, with the most pronounced improvement observed in VisualSearch (+23.8 points). 
Notably, even without any fine-tuning, simply equipping strong off-the-shelf models with our toolset provides significant benefits: GPT-5.5 and Kimi K2.6 achieve 7.1 and 14.5 absolute gains, respectively, demonstrating the inherent utility of the designed tools.
However, smaller open-source models derive little or even negative benefit from tool access without dedicated training. Fine-tuning on \dataset enables them to use tools effectively, bringing their performance close to that of closed-source models. For example, Qwen3.5-9B with our training achieves an average of 45.8, surpassing GPT-5.5's 41.9 without tools and closely approaching its tool-augmented performance of 49.0.
In contrast, existing tool-use models Thyme and DeepEyes V2, trained on narrow single-domain data, perform worse on the five-domain average than vanilla base models \emph{without any tool access}—they improve on their specialty but collapse catastrophically elsewhere. This comparison highlights that a diverse, multi-domain training corpus, as provided by \dataset{}, is essential for robust and generalizable visual tool use.

\paragraph{Largest gains emerge where tools unlock new visual evidence.}
For domains where the answer is largely derivable from surface-level visual content—such as Chart, which primarily tests text and semantic extraction from images—a sufficiently capable model can already perform well without tools (e.g., GPT-5.5 scores 63.3 on Chart without tools). However, for more challenging domains involving dense details and hidden evidence—such as VisualSearch, which demands fine-grained spatial and geometric reasoning—models often struggle to succeed with their inherent visual perception alone. In these scenarios, invoking auxiliary tools to crop, zoom, or reformat visual content becomes essential. This explains why our method yields the most substantial gains precisely in these demanding domains (e.g., Qwen3.5-9B: +23.8 points on VisualSearch), where tools effectively bridge the gap between what the model can perceive and what the task requires.

\subsection{Out-of-Distribution Generalization}
\label{sec:ood}

To test whether the tool-use capability learned from our five training domains transfers beyond them, we further evaluate on Agentic-MME and VTC-Bench with out-of-distribution domains held out from training. Table~\ref{tab:ood} shows that, across all four backbones, models trained on \dataset{} consistently outperform their corresponding no-tool backbones on both Agentic-MME and VTC-Bench. This consistent improvement on benchmarks outside the five training domains suggests that \dataset{} teaches transferable visual evidence-acquisition behavior rather than merely encouraging domain-specific tool-calling patterns.

\input{tables/tab_cross_domain}

\subsection{Ablation Study}
\label{sec:ablation_gain}
\label{sec:notool_traj}
To validate that effective visual tool-use learning requires both correct answers and causally useful tool traces, we compare four Qwen3.5-9B supervision variants under the same training budget. The first three use tool trajectories retained by (1)~both outcome validity and causal utility (full \dataset{}), (2)~outcome validity alone, or (3)~causal utility alone. The fourth is a no-tool-trajectory SFT baseline, constructed by re-synthesizing trajectories without tool invocation for the queries in the full variant.

\paragraph{Gains originate from tool-use, not from SFT alone.}
As shown in Figure~\ref{fig:ablation}, every tool-augmented variant outperforms the no-tool-trajectory SFT baseline, confirming that improvements stem from learned tool invocation rather than simply training on additional difficult queries.

\paragraph{Both filters are complementary and necessary.}
Intersecting both filters consistently dominates either filter in isolation, with outcome-validity-only and causal-utility-only trailing by 1.7 and 3.4 points, respectively.
The two filters address distinct failure modes: outcome validity removes demonstrations that lead to incorrect answers and would teach erroneous behavior, while causal utility discards trajectories where tools are invoked but their observations contribute no measurable benefit to reaching the solution. Their combination yields the highest-quality training signal.

\paragraph{Instructive filtering yields more useful tool interactions.}
% To probe the learned behavior beyond the accuracy of the fine-tuned models themselves, we replay their tool-interaction trajectories to the original Qwen3.5-9B probe model. The replayed context contains only tool call arguments and returned observations; all reasoning and final answers from the trajectory-generating model are excluded, ruling out direct copying of the generator's answer or imitation of its rationale.
To probe the learned behavior beyond the task accuracy of the fine-tuned models themselves, we conduct a trajectory replay analysis using the original Qwen3.5-9B probe model. Specifically, we replay the tool-interaction trajectories generated by fine-tuned models. The replayed context includes only the tool call arguments and their corresponding returned observations. We deliberately exclude all intermediate reasoning and final answers, which prevents the probe model from directly copying the generator’s answer or simply imitating its rationale.
As shown in Table~\ref{tab:trajectory_replay}, trajectories from both trained models improve the probe model, but those from the model trained on instructive trajectories consistently outperform correctness-only trajectories. This behavior-level transfer suggests that causal-utility filtering induces an evidence-acquisition policy whose resulting interactions are more useful to another model, rather than merely producing tool calls that accompany successful outcomes.

\subsection{Cross-Domain Transfer}
\label{sec:cross_domain}

% \paragraph{Setup.}
We fine-tune Qwen3.5-9B on each single-domain slice of \dataset{} separately and evaluate on all five domains. Full results are reported in Table~\ref{tab:cross}.

\paragraph{Tool-use transfers as a domain-general meta-skill.}
Training on any single domain improves overall performance over the tool-enabled base model, with gains often extending to domains unseen during training. Most notably, GUI-only training improves VisualSearch by 14.1 points. This cross-domain benefit suggests that the model learns reusable tool-use primitives, such as localized cropping and spatial grounding, rather than only domain-specific solutions.

\paragraph{Transfer is asymmetric, with both positive and negative effects.}
The transfer is not uniformly beneficial because different domains reward different interaction patterns. For example, VisualSearch-only training lowers Web2HTML performance from 42.0 to 21.6. We attribute this degradation to a mismatch in tool-use strategies: repeatedly cropping small regions is effective for fine-grained search but conflicts with  global, full-page reasoning. Single-domain training can therefore over-specialize the tool-use policy and harm domains that require incompatible strategies.

\paragraph{The full mixture resolves strategy conflicts.}
The full \dataset{} mixture achieves a 45.8 average, surpassing all single-domain variants, and performs best on four of the five domains. Exposure to diverse interaction patterns helps the model resolve strategy conflicts through \emph{context-conditional} tool use: it learns not only how to invoke tools, but also which strategy is appropriate for the visual input. GUI is the only exception: its fixed localization protocol and point-in-box metric favor specialized coordinate calibration, allowing GUI-only training to outperform the full mixture. Nevertheless, the full mixture still substantially improves over the tool-enabled base model, demonstrating effective cross-domain transfer.

% \paragraph{The GUI exception.}
% GUI is the only domain where a specialist trained on its single-domain slice outperforms the full mixture on the corresponding domain. A likely reason is that the point-in-box metric rewards a single precise click, making direct imitation especially effective, whereas other domains emphasize multi-step interactions such as crop$\to$OCR. Because the full mixture still improves GUI over the untuned tool-enabled base model, this exception points to domain-adaptive mixture weighting rather than a failure of multi-domain tool-use supervision.

%% file: tables/tab_main.tex
\begin{table*}[t]
\centering
% \small

% Keep the compact column spacing local to the tabular.
\begingroup
\setlength{\tabcolsep}{3pt}

\begin{tabularx}{\textwidth}{@{}ll*{6}{Y}@{}}

\toprule
\multirow{2}{*}{\textbf{Model}}
& \multirow{2}{*}{\textbf{Setting}}
& \multicolumn{6}{c}{\textbf{OpenVisTool-Bench}}
\\
\cmidrule(lr){3-8}

& & \textbf{Chart}
& \textbf{GUI}
& \textbf{Table}
& \textbf{VisualSearch}
& \textbf{Web2HTML}
& \textbf{AVG} \\
\midrule

\multirow{2}{*}{GPT-5.5}
& w/o tool
& 63.3
& 17.7
& 48.9
& 27.7
& 51.7
& 41.9 \\

& with tool
& \textbf{66.6}
& 24.2
& \textbf{53.9}
& 46.2
& \textbf{54.3}
& \underline{49.0} \\

\addlinespace[1pt]

\multirow{2}{*}{Kimi K2.6}
& w/o tool
& 50.9
& 17.1
& 34.4
& 13.8
& 49.4
& 33.1 \\

& with tool
& \underline{64.3}
& 18.8
& \underline{51.6}
& 49.8
& \underline{53.6}
& 47.6 \\

\midrule
Qwen2.5-VL-7B
& w/o tool
& 18.2
& 5.1
& 21.2
& 28.3
& 23.1
& 19.2 \\

\addlinespace[1pt]

Thyme
& with tool
& 19.8
& 0.0
& 17.9
& 39.2
& 14.3
& 18.2 \\

\addlinespace[1pt]

DeepEyes V2
& with tool
& 20.9
& 1.7
& 20.4
& 31.8
& 2.4
& 15.5 \\

\midrule
\multirow{2}{*}{Qwen3.5-4B}
& w/o tool
& 31.6
& 22.0
& 25.4
& 34.0
& 40.8
& 30.8 \\

& with tool
& 32.5
& 23.5
& 32.3
& 39.6
& 41.6
& 33.9 \\

\quad+\dataset
& with tool
& 43.0\imp{11.4}
& 27.4\imp{5.4}
& 37.3\imp{11.9}
& 56.6\imp{22.6}
& 43.4\imp{2.6}
& 41.5\imp{10.7} \\

\midrule
\multirow{2}{*}{Qwen3-VL-8B-Instruct}
& w/o tool
& 22.1
& 13.0
& 26.7
& 31.8
& 26.5
& 24.0 \\

& with tool
& 20.9
& 12.6
& 24.6
& 30.2
& 28.1
& 23.3 \\

\quad+\dataset
& with tool
& 26.1\imp{4.0}
& 19.7\imp{6.7}
& 35.3\imp{8.6}
& 50.7\imp{18.9}
& 32.9\imp{6.4}
& 32.9\imp{8.9} \\

\midrule
\multirow{2}{*}{Qwen3.5-9B}
& w/o tool
& 34.6
& 27.1
& 29.0
& 35.6
& 40.1
& 33.3 \\

& with tool
& 28.4
& 24.8
& 26.8
& 38.7
& 42.0
& 32.1 \\

\quad+\dataset
& with tool
& 49.3\imp{14.7}
& \textbf{31.6}\imp{4.5}
& 43.1\imp{14.1}
& \textbf{59.4}\imp{23.8}
& 45.5\imp{5.4}
& 45.8\imp{12.5} \\

\midrule
\multirow{2}{*}{Qwen3.5-27B}
& w/o tool
& 45.7
& 28.9
& 29.8
& 43.2
& 46.6
& 38.8 \\

& with tool
& 47.7
& \underline{29.1}
& 43.3
& 54.7
& 44.3
& 43.8 \\

\quad+\dataset
& with tool
& 60.7\imp{15.0}
& \textbf{31.6}\imp{2.7}
& 48.2\imp{18.4}
& \underline{57.3}\imp{14.1}
& 48.7\imp{2.1}
& \textbf{49.3}\imp{10.5} \\

\bottomrule
\end{tabularx}
\endgroup

\caption{
Main results on OpenVisTool-Bench under the avg@4 evaluation protocol. ``w/o tool'' and ``with tool'' denote evaluation without and with access to the corresponding visual tool environment, respectively. Rows marked with +\dataset denote models fine-tuned on our training data, and green $+x$ suffixes report absolute gains over the ``w/o tool'' result of the same backbone. \textbf{Bold} and \underline{underlined} entries indicate the best and second-best results in each column, respectively.
}
\label{tab:main}

\end{table*}

%% file: sections/bench.tex
\subsection{\benchname: Isolating Tool-Use Ability}
\label{sec:bench}

General-purpose visual benchmarks often mix tool-relevant instances with questions that can already be solved from the initial image encoding. \benchname instead isolates the incremental value of active visual evidence acquisition by focusing on instances where operations such as cropping, enhancement, or structure detection materially improve task performance. Whereas Agentic-MME~\citep{wei2026agentic} evaluates broad multimodal agency and VTC-Bench~\citep{zhu2026vtc} stresses compositional tool execution, \benchname specifically measures the performance gain enabled by visual-tool access.

\paragraph{Construction.}
We construct each domain according to whether its source benchmark already targets tool-demanding tasks.

\begin{itemize}

\item \textbf{Chart and Table.}
We source Chart instances from CharXiv \citep{charxiv} and ChartMuseum \citep{chartmuseum}, and Table instances from TableVQA-Bench \citep{tablevqabench} and MMTBench \citep{mmtbench}. For each instance, three strong models (GPT-5.4, Gemini-3.0-Flash, and Qwen3.5-Plus) are evaluated with and without tools using five trials per condition. We retain the instance if at least one model achieves a tool-use gain of $\mathrm{avg@5}_{\mathrm{tool}}-\mathrm{avg@5}_{\mathrm{no\text{-}tool}} \geq 0.4$.

\item \textbf{GUI Grounding, Visual Search, and Web-to-HTML.}
These sources already require active visual interaction---precise localization, fine-grained retrieval, and iterative rendering---so we apply no additional tool-gain filtering. We use the 117 ScreenSpot-Pro~\citep{li2025screenspotpro} instances with the smallest target regions, together with VisualProbe-Hard~\citep{lai2026minio3} and Vision2Web-Level1~\citep{he2026vision2web}.

\end{itemize}

More details on the construction of \benchname, including source-specific sampling and filtering procedures, are provided in Appendix E.

\paragraph{Evaluation protocol.}
Each domain uses a task-specific evaluator. Chart, Table, and Visual Search use LLM-as-a-judge for binary correctness. GUI uses rule-based point-in-box accuracy. Web-to-HTML follows the official Vision2Web evaluation protocol, measuring component-level visual fidelity between the rendered output and reference page on a $0$--$100$ scale. We report each domain score and the macro-average across domains under the avg@4 protocol.

%% file: tables/tab_ood.tex
\begin{table}[t]
\centering

\begingroup
\small
\setlength{\tabcolsep}{4pt}
\begin{tabular*}{\columnwidth}{@{\extracolsep{\fill}}lcc@{}}

\toprule
Model
& Agentic-MME
& VTC-Bench \\
\midrule

Qwen3.5-4B
& 29.9
& 27.5 \\

\quad+\dataset
& 34.5\imp{4.6}
& 38.8\imp{11.3} \\

\midrule
Qwen3-VL-8B-Instruct
& 25.6
& 28.2 \\

\quad+\dataset
& 32.3\imp{6.7}
& 29.1\imp{0.9} \\

\midrule
Qwen3.5-9B
& 32.7
& 30.1 \\

\quad+\dataset
& 41.6\imp{8.9}
& 41.4\imp{11.3} \\

\midrule
Qwen3.5-27B
& 37.1
& 34.9 \\

\quad+\dataset
& 41.1\imp{4.0}
& 45.2\imp{10.3} \\

\bottomrule
\end{tabular*}
\endgroup

\caption{
Out-of-distribution results on Agentic-MME and VTC-Bench. Base models are evaluated without tools, while models trained on OpenVisTool-42K are evaluated with tools. Green suffixes show absolute gains over the corresponding no-tool backbone.
}
\label{tab:ood}

\end{table}

%% file: tables/tab_trajectory_replay.tex
\begin{table}[t]
    \centering
    \small
    \setlength{\tabcolsep}{3.5pt}
    \begin{tabular}{@{}lccc@{}}
    \toprule
    \textbf{Trajectory source} & \textbf{\shortstack{OpenVisTool-\\Bench}} & \textbf{\shortstack{Agentic-\\MME}} & \textbf{\shortstack{VTC-\\Bench}} \\
    \midrule
    None (Qwen3.5-9B w/o tool) & 33.3 & 32.7 & 30.1 \\
    Correctness-only model & 42.1 & 36.5 & 36.5 \\
    Instructive model (ours) & \textbf{44.5} & \textbf{37.7} & \textbf{38.1} \\
    \bottomrule
    \end{tabular}
    \caption{Trajectory replay analysis using the original Qwen3.5-9B probe model. The probe model answers each query conditioned on the tool call arguments and returned observations generated by the correctness-only or instructively trained model. The generator's reasoning and final answer are excluded. ``None'' denotes evaluation without tool access or trajectory replay.}
    \label{tab:trajectory_replay}
\end{table}

%% file: tables/tab_cross_domain.tex
\begin{table*}[t]
\centering
\small

\begin{tabular}{llcccccc}
\toprule
\textbf{Variant} & \textbf{Setting} & \textbf{Chart} & \textbf{GUI} & \textbf{Table} & \textbf{VisualSearch} & \textbf{Web2HTML} & \textbf{AVG} \\
\midrule
\multirow{2}{*}{Qwen3.5-9B} & w/o tool & 34.6 & 27.1 & 29.0 & 35.6 & 40.1 & 33.3 \\
 & with tool & 28.4 & 24.8 & 26.8 & 38.7 & 42.0 & 32.1 \\
\midrule
+ \dataset{} (full) & with tool & \textbf{49.3} & \underline{31.6} & \textbf{43.1} & \textbf{59.4} & \textbf{45.5} & \textbf{45.8} \\
\midrule
\multicolumn{8}{l}{\textit{Single-domain slices}} \\
\quad + Chart & with tool & \underline{46.6} & 22.2 & 37.1 & 44.3 & 36.5 & 37.4 \\
\quad + GUI & with tool & 41.4 & \textbf{36.3} & 38.7 & 52.8 & 39.1 & \underline{41.7} \\
\quad + Table & with tool & 45.7 & 20.5 & 38.9 & 45.8 & 39.2 & 38.0 \\
\quad + VisualSearch & with tool & 45.7 & 23.3 & \underline{40.9} & \underline{57.3} & 21.6 & 37.8 \\
\quad + Web2HTML & with tool & 41.6 & 18.2 & 38.5 & 42.2 & \underline{44.6} & 37.0 \\
\bottomrule
\end{tabular}

\caption{Cross-domain transfer analysis on \benchname with Qwen3.5-9B as base model. We fine-tune the model on either the full \dataset{} mixture or an individual domain slice and evaluate all fine-tuned variants with tools across the five domains. \textbf{Bold} and \underline{underlined} entries indicate the best and second-best results in each column, respectively.}
\label{tab:cross}

\end{table*}

%% file: sections/conclusion.tex
\section{Conclusion}
\label{sec:conclusion}
In this work, we introduce \method, a framework for constructing \textbf{instructive} visual tool-use trajectories that addresses a limitation of outcome-only filtering: an answer-correct trajectory may contain tool observations that do not contribute to its answer. \method defines instructive supervision through outcome validity and causal utility, and operationalizes these criteria with difficulty screening, domain-specific trajectory synthesis, and supervision verification to construct \dataset across five domains. We separately build \benchname across the same domains to evaluate models' ability to acquire and use visual evidence through tools. Across four backbones, training on \dataset consistently improves performance. Trajectory replay and cross-domain results further suggest that instructive supervision promotes useful evidence acquisition and transferable tool-use behavior.

%% file: sections/appendix.tex
% =========== 按照正文出现的顺序 ===========
% A. OpenVisTool-42K Construction
% B. Complete Toolset
% C. Domain-Specific Instructions
% D. Qualitative Trajectories
% E. OpenVisTool-Bench Construction
% F. Additional Experimental Results
% G. Implementation Details
% =======================================

\begin{figure*}[t]
\centering
\includegraphics[width=0.49\textwidth]{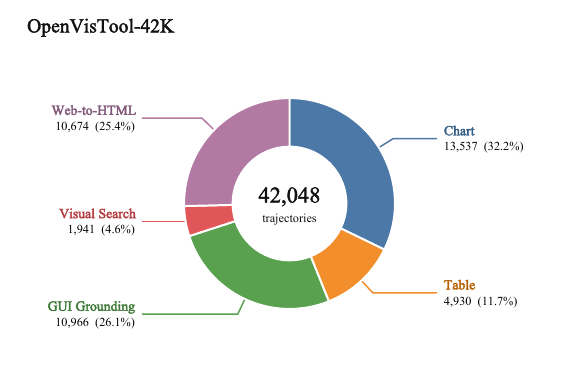}
\hfill
\includegraphics[width=0.49\textwidth]{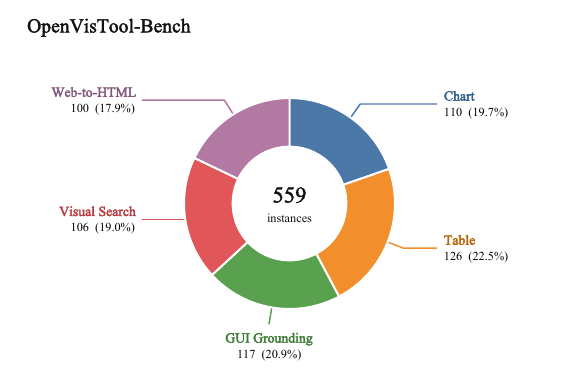}
\caption{\textbf{Domain distributions of OpenVisTool-42K and OpenVisTool-Bench.} The 42,048 retained training trajectories reflect the natural yield of the shared filtering pipeline (left), while the 559 evaluation instances are approximately balanced across the same five domains (right).}
\label{fig:domain_distributions}
\end{figure*}

\section{OpenVisTool-42K Construction Details}
\label{app:data_detail}

\subsection{Source Datasets and Preprocessing}

For each example, the task image and query serve as the task input to the teacher, whose rollout is additionally conditioned on the domain-specific tool-use instructions detailed in Section~\ref{app:prompts}. The reference answer is withheld from the teacher and used only offline for source preprocessing where applicable, difficulty screening, and outcome-validity verification. We do not use source-provided reasoning traces or tool-use trajectories as supervision. We first apply the domain-specific preprocessing described below, after which all candidate pools undergo the uniform difficulty screening introduced in Section~3.2 of the main paper.

\paragraph{Chart.}
We use 100,000 examples from ChartVerse-SFT \citep{chartverse} without additional preprocessing.

\paragraph{Table.}
We draw from CoSyn \citep{cosyn} and TABLET \citep{tablet}. From CoSyn, we select the table-image subset and discard examples with an empty or invalid question or answer. From TABLET, we use examples from the HiTab, TabMWP, TAT-QA, and WikiTQ training subsets. After preprocessing, the Table candidate pool contains 469,460 examples.

\paragraph{GUI Grounding.}
We draw from OS-Atlas \citep{osatlas}, AgentNet \citep{agentnet}, and UGround \citep{uground}. From the Linux, macOS, and Windows portions of OS-Atlas, we discard annotations with missing instructions or images, malformed or non-normalized target boxes, low-resolution images ($\leq 1$ MP), or boxes covering at least 0.5\% of the image. From AgentNet, we retain the high-resolution Ubuntu click subset. From UGround, we discard examples with empty, URL-like, or overly short instructions, invalid or overly large target boxes, low-resolution images, or non-landscape layouts, and then subsample the filtered pool. After preprocessing, the GUI Grounding candidate pool contains 698,422 examples.

\paragraph{Visual Search.}
We draw from Vero-600K \citep{vero} and the DeepEyesV2-RL corpus \citep{hong2026deepeyesv2}. From Vero-600K, we select the PixelReasoner and VisualProbe components, yielding 9,744 examples. From DeepEyesV2-RL, we retain images whose shorter side is at least 768 pixels and remove chart-like examples using keyword rules followed by classification with Qwen3-VL-30B-A3B-Instruct, yielding 9,285 examples. The resulting Visual Search candidate pool contains 19,029 examples.

\paragraph{Web-to-HTML.}
We draw from VinciCoder \citep{vincicoder}. We remove screenshots dominated by a non-background color and construct a diverse pool spanning complex UIs, high-pixel-count screenshots, and typical 1280$\times$720 layouts. We also replace the heterogeneous source prompts with a shared screenshot-to-HTML reconstruction instruction before teacher rollout. After preprocessing, the Web-to-HTML candidate pool contains 26,829 examples.

\subsection{Trajectory Synthesis and Filtering Details}
\label{app:synthesis_config}

\paragraph{Teacher rollout.}
We synthesize trajectories with Qwen3.5-Plus in a function-calling agent loop. At each turn, the teacher produces a reasoning step and optional tool calls; each returned observation is appended to the context. Domain-specific instructions are injected through the rollout configuration, while all domains share the same toolset. Each query runs in an isolated workspace with its own media directory, preventing generated crops, annotations, masks, and rendered pages from colliding across examples. The loop terminates when the teacher returns a final answer or, for GUI Grounding, emits the terminal \texttt{computer\_use} click.

\paragraph{Outcome validity and trajectory sanitation.}
After rollout, we first evaluate the teacher's final output. Chart uses rule-based answer matching; Table and Visual Search use an LLM judge (Qwen3.5-27B); GUI Grounding uses point-in-box evaluation; and Web-to-HTML uses rendered-page comparison with a VLM judge. For Web-to-HTML, the generated HTML must be extractable and renderable and must receive a visual-consistency score of at least 80 from Qwen3.5-27B. We then reject trajectories with no tool call, no valid final response (or no terminal coordinate for GUI Grounding), more than 30 tool rounds, missing observations or generated images, failed tool executions, or invalid file references. Only trajectories that pass both outcome evaluation and these sanitation checks proceed to causal-utility filtering.

\paragraph{Causal-utility filtering.}
For each surviving trajectory, we provide the original image and query together with the recorded tool calls and responses, in chronological order, to the same Qwen3.5-9B probe used for difficulty screening. Generated images returned by tools are included with their corresponding responses, whereas the teacher's reasoning and final answer are excluded to prevent answer leakage. We rerun the probe four times to compute the trajectory-conditioned avg@4 and compare it with the corresponding no-tool avg@4. We retain the trajectory when the improvement is at least $0.1$; together with the preceding outcome filter, this ensures that every retained trajectory satisfies both outcome validity and causal utility. Before constructing the context, each serialized tool call and its corresponding response are independently limited to 10,000 characters. The maximum generation lengths are 32,768 tokens for both difficulty screening and the causal-utility probe.

\subsection{Final Dataset Statistics and Domain Distribution}

The final training corpus contains 42,048 trajectories and 296,993 executed tool calls, averaging 7.06 calls per trajectory; every retained trajectory contains at least one tool invocation. Figure~\ref{fig:domain_distributions}  (left) visualizes the domain composition. The distribution reflects the natural yield after the shared difficulty, outcome-validity, and causal-utility filters, without domain-level upsampling or rebalancing.

\section{The Complete Toolset}
\label{app:toolset}

Our shared toolset consists of general-purpose tools and visual tools, summarized in Table~\ref{tab:tools-complete}.

\paragraph{General-Purpose Tools.}
The general-purpose tools provide file and system operations shared across all task domains.

\paragraph{Visual Tools.}
The visual tools are grouped by functionality. For GUI Grounding, we additionally use \texttt{computer\_use} to express the final GUI action (e.g., \texttt{click}) in the required format without executing it.

\section{Domain-Specific Tool-use Instructions}
\label{app:prompts}

Tables~\ref{tab:prompt-chart}--\ref{tab:prompt-web-to-html} provide the domain-specific instructions injected into the teacher's system prompt during trajectory rollout.

\section{Qualitative Trajectory Examples}
\label{app:qualitative}

Figures~\ref{fig:case-chart-retained}--\ref{fig:case-rejected} contrast two retained trajectories with an outcome-correct trajectory rejected by causal utility.  We show only the decision-relevant reasoning and tool observations; the reported $\bar p_0$ and $\bar p_{\mathrm{tool}}$ are the corresponding four-trial probe success rates (avg@4).

\tcbset{
    ovtcase/.style={
    enhanced,
    boxrule=0.55pt,
    arc=1.2mm,
    left=2mm,
    right=2mm,
    top=1.4mm,
    bottom=1.4mm,
    before skip=0pt,
    after skip=2mm
    },
    ovttrace/.style={
    enhanced,
    colback=black!2,
    colframe=black!25,
    boxrule=0.4pt,
    arc=0.8mm,
    left=1.6mm,
    right=1.6mm,
    top=1mm,
    bottom=1mm,
    before skip=1mm,
    after skip=1mm
    }
}

\providecommand{\ovttool}[1]{%
    \tcbox[
    on line,
    colback=blue!5,
    colframe=blue!35!black,
    boxrule=0.35pt,
    arc=0.7mm,
    left=0.7mm,
    right=0.7mm,
    top=0.25mm,
    bottom=0.25mm
    ]{\ttfamily\scriptsize #1}%
}

\section{OpenVisTool-Bench Construction Details}
\label{app:bench_detail}

OpenVisTool-Bench contains 559 independently curated instances spanning the same five domains as the training corpus. Its domain composition is shown in the Figure~\ref{fig:domain_distributions} (right).

\paragraph{Chart and Table.}
The Chart source pool consists of 1,000 instances from the CharXiv validation set~\citep{charxiv} and 1,162 from ChartMuseum~\citep{chartmuseum}, while the Table source pool consists of 1,500 instances from TableVQA-Bench~\citep{tablevqabench} and 4,022 from MMTBench~\citep{mmtbench}.  For both domains, we screen instances with GPT-5.4, Gemini-3.0-Flash, and Qwen3.5-Plus, running five trials with tools and five without tools.  We retain the cross-model union of instances for which at least one model achieves $\mathrm{avg@5}_{\mathrm{tool}}-\mathrm{avg@5}_{\mathrm{no\text{-}tool}}>0.4$.  This yields 110 of the 2,162 Chart instances and, after restricting the TableVQA-Bench contribution to its VWTQ and VWTQ-Syn subsets, 126 of the 5,522 Table instances.

\paragraph{GUI Grounding.}
ScreenSpot-Pro~\citep{li2025screenspotpro} contains 1,581 instances.  We retain the 117 instances whose relative target-box area falls between $5.70\!\times\!10^{-5}$ and $8.43\!\times\!10^{-5}$ of the screenshot area, without model-performance filtering.

\paragraph{Visual Search.}
We directly include all 106 instances from VisualProbe-Hard~\citep{lai2026minio3}, which targets fine-grained exploratory visual search, without additional filtering.

\paragraph{Web-to-HTML.}
We use all 100 Level-1 static-webpage tasks from Vision2Web~\citep{he2026vision2web}, where an agent reconstructs a responsive page from desktop, tablet, and mobile visual prototypes, without additional filtering.

\begin{table*}[t]
\centering
\small
\setlength{\tabcolsep}{7pt}
\renewcommand{\arraystretch}{1.08}
\begin{tabular*}{0.8\textwidth}{@{\extracolsep{\fill}}llcccc}
\toprule
\textbf{Model} & \textbf{Setting} &
\multicolumn{2}{c}{\textbf{Chart}} &
\multicolumn{2}{c}{\textbf{Table}} \\
\cmidrule(lr){3-4}\cmidrule(lr){5-6}
& & \textbf{CharXiv} & \textbf{ChartMuseum} &
\textbf{TableVQA-Bench} & \textbf{MMTBench} \\
\midrule
Qwen3.5-4B
    & w/o tool  & 63.7 & 48.7 & 70.7 & 32.7 \\
\quad + \dataset{}
    & with tool & 65.5 & 57.8 & 86.7 & 36.5 \\
\addlinespace[2pt]
Qwen3.5-9B
    & w/o tool  & 67.2 & 61.2 & 82.1 & 36.5 \\
\quad + \dataset{}
    & with tool & 68.3 & 66.9 & 88.5 & 39.6 \\
\addlinespace[2pt]
Qwen3-VL-8B-Instruct
    & w/o tool  & 51.2 & 40.4 & 83.3 & 33.3 \\
\quad + \dataset{}
    & with tool & 54.0 & 46.7 & 84.9 & 36.1 \\
\bottomrule
\end{tabular*}
\caption{Results on the complete, unfiltered Chart and Table source benchmarks.  The setting indicates whether the model is trained with tool-use trajectories; tool invocation is disabled at inference time for all models.}
\label{tab:fullset}
\end{table*}

\begin{table}[t]
\centering
\small
\setlength{\tabcolsep}{4pt}
\begin{tabularx}{\columnwidth}{@{}p{0.44\columnwidth}X@{}}
\toprule
\textbf{Hyperparameter} & \textbf{Setting} \\
\midrule
Epochs & 3 \\
Optimizer steps & 1,971 \\
Global batch size & 64 \\
Optimizer & Adam \\
Adam $\beta_1,\beta_2$ & 0.9, 0.95 \\
Learning rate & $1\!\times\!10^{-5}$ \\
LR schedule & Cosine decay \\
Warmup ratio & 0.03 \\
Weight decay & 0.1 \\
Gradient clipping & 1.0 \\
Maximum sequence length & 65,536 tokens \\
Maximum image tokens & 8,192 \\
\bottomrule
\end{tabularx}
\caption{Shared training settings for all four backbones.}
\label{tab:training-settings}
\end{table}

\begin{table}[t]
\centering
\small
\setlength{\tabcolsep}{5pt}
\begin{tabularx}{\linewidth}{@{}p{0.46\linewidth}X@{}}
\toprule
\textbf{Hyperparameter} & \textbf{Setting} \\
\midrule
Temperature & 0.6 \\
Top-$p$ & 0.95 \\
Top-$k$ & 20 \\
Presence penalty & 0.0 \\
Repetition penalty & 1.0 \\
Maximum output length & 32,768 \\
Maximum agent turns & 50 \\
Judge model & GPT-5.5 \\
\bottomrule
\end{tabularx}
\caption{Evaluation and inference settings.}
\label{tab:evaluation-settings}
\end{table}

\section{Additional Experimental Results}
\label{app:fullset}

This section provides the full-benchmark evaluation referenced in the Experimental Setup of the main paper.  Because \benchname{} uses challenging subsets drawn from the Chart and Table source benchmarks, we additionally evaluate each backbone and its counterpart trained on \dataset{} on the complete, unfiltered source test sets.  We disable tool invocation for this evaluation to isolate the capabilities transferred to the models themselves.

As shown in Table~\ref{tab:fullset}, training on \dataset{} improves every backbone on every full source benchmark.  The pattern is consistent across both chart understanding and table reasoning, rather than being concentrated in one dataset or model family.  The gains are particularly clear on ChartMuseum across backbones, while the smaller Qwen3.5 model also benefits substantially on TableVQA-Bench; configurations that begin from stronger baselines generally show more moderate but still reliable improvements. Their consistency on the unfiltered source benchmarks supports that the improvements on \benchname{} are not an artifact of subset selection.

\section{Implementation Details}
\label{app:implement}

\paragraph{Training.}
We fine-tune all backbones with SWIFT~\citep{zhao2025swift}. Throughout training, we freeze the vision encoder and merger and update only the LLM parameters. The training hyperparameters are summarized in Table~\ref{tab:training-settings}.

\paragraph{Evaluation.}
We serve the evaluated open-source models with the vLLM backend. We use the same decoding configuration for the base and fine-tuned Qwen models, as listed in Table~\ref{tab:evaluation-settings}. GPT-5.5 is the judge model for all judge-based tasks. GUI Grounding is scored deterministically by whether the predicted click falls inside the ground-truth bounding box.

% Collect all display-heavy material after the continuous A--G narrative.
\input{tables/tab_tool_complete}
\input{tables/tab_domain_prompts}
% Provenance: ChartVerse id 51236; session_51236.
% p0=.50, ptool=.75, correctness=bold_approx.
\begin{figure*}[p]
\centering
\small

\begin{tcolorbox}[
    ovtcase,
    colback=green!3,
    colframe=green!45!black
]
\textbf{\textsc{Retained} $\boldsymbol{\cdot}$ Chart: decompose stacked bars by color}
\hfill
{\footnotesize $\bar p_0=0.50 \;\longrightarrow\; \bar p_{\mathrm{tool}}=0.75$, \textbf{$g=+0.25$}}

\smallskip
\textbf{Question.} What is the average ratio of high-calorie consumption to combined low- and moderate-calorie consumption across all food categories?
\end{tcolorbox}

\begin{tabularx}{0.92\textwidth}{@{}>{\centering\arraybackslash}X@{}}
\includegraphics[width=\linewidth]{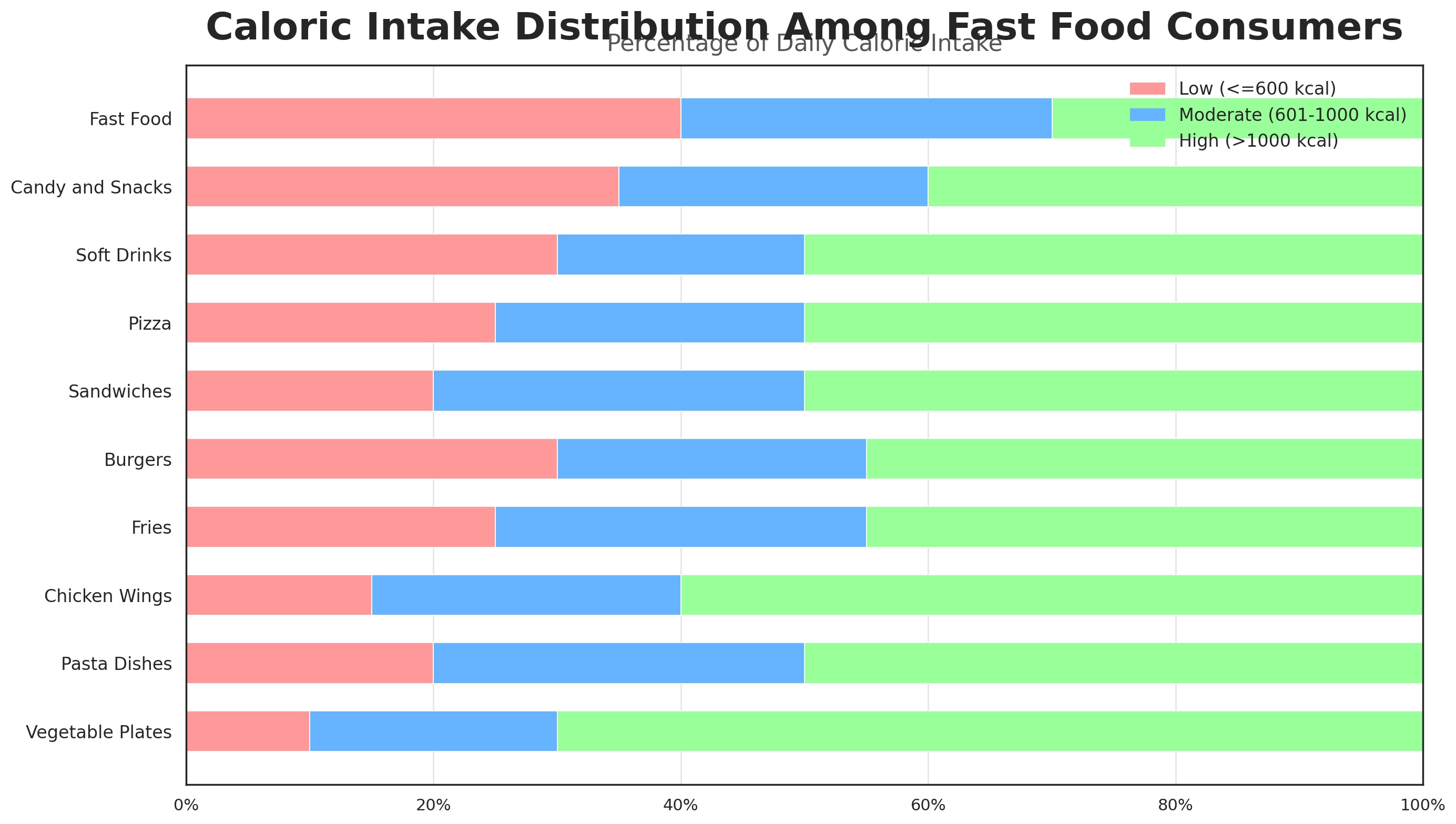}\\[-1pt]
{\scriptsize (a) Input: each category contains three adjacent segments whose widths must be measured consistently.}
\end{tabularx}

\vspace{1.5mm}

\begin{tabularx}{\textwidth}{@{}*{3}{>{\centering\arraybackslash}X}@{}}
\includegraphics[width=\linewidth]{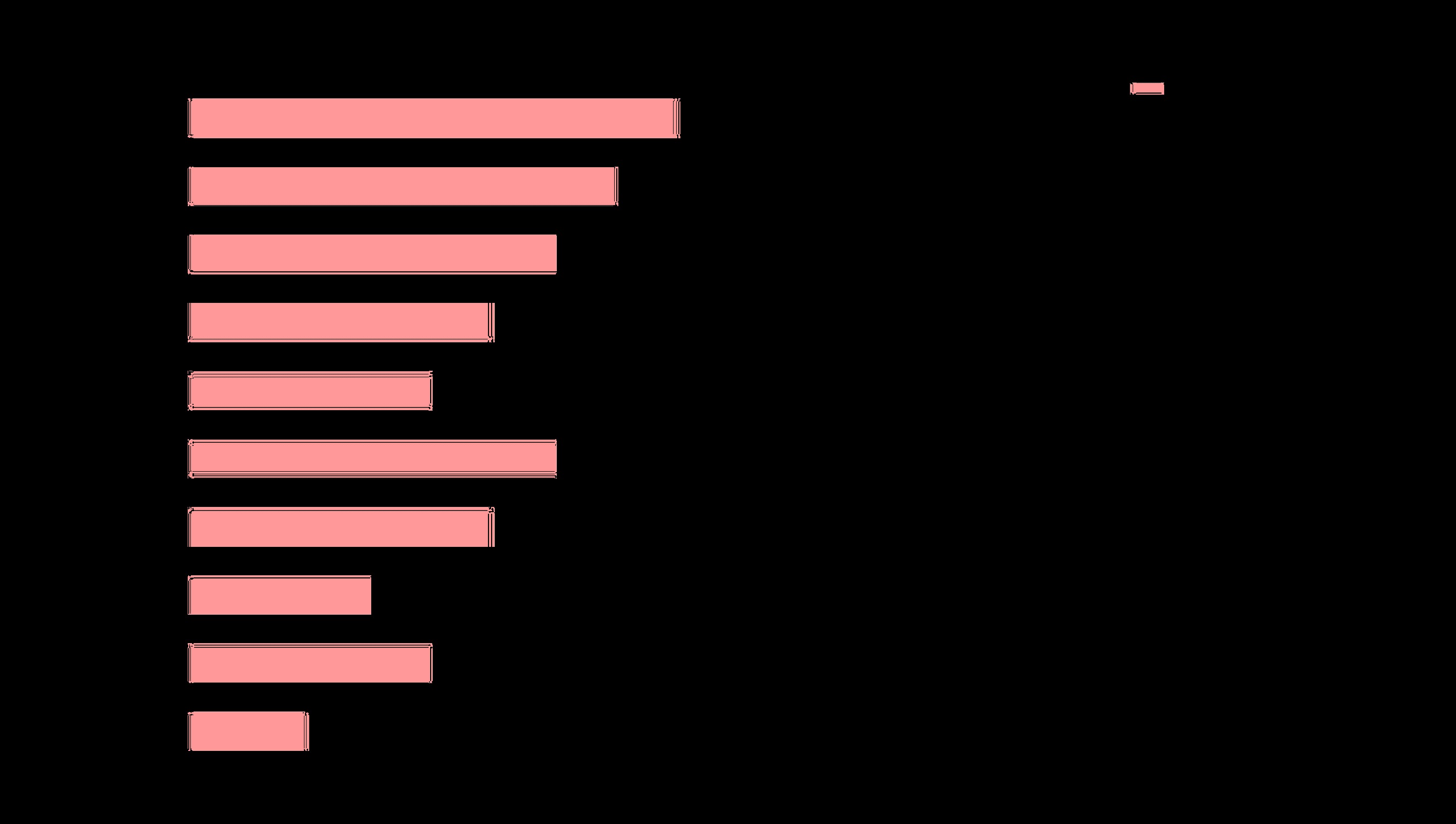} &
\includegraphics[width=\linewidth]{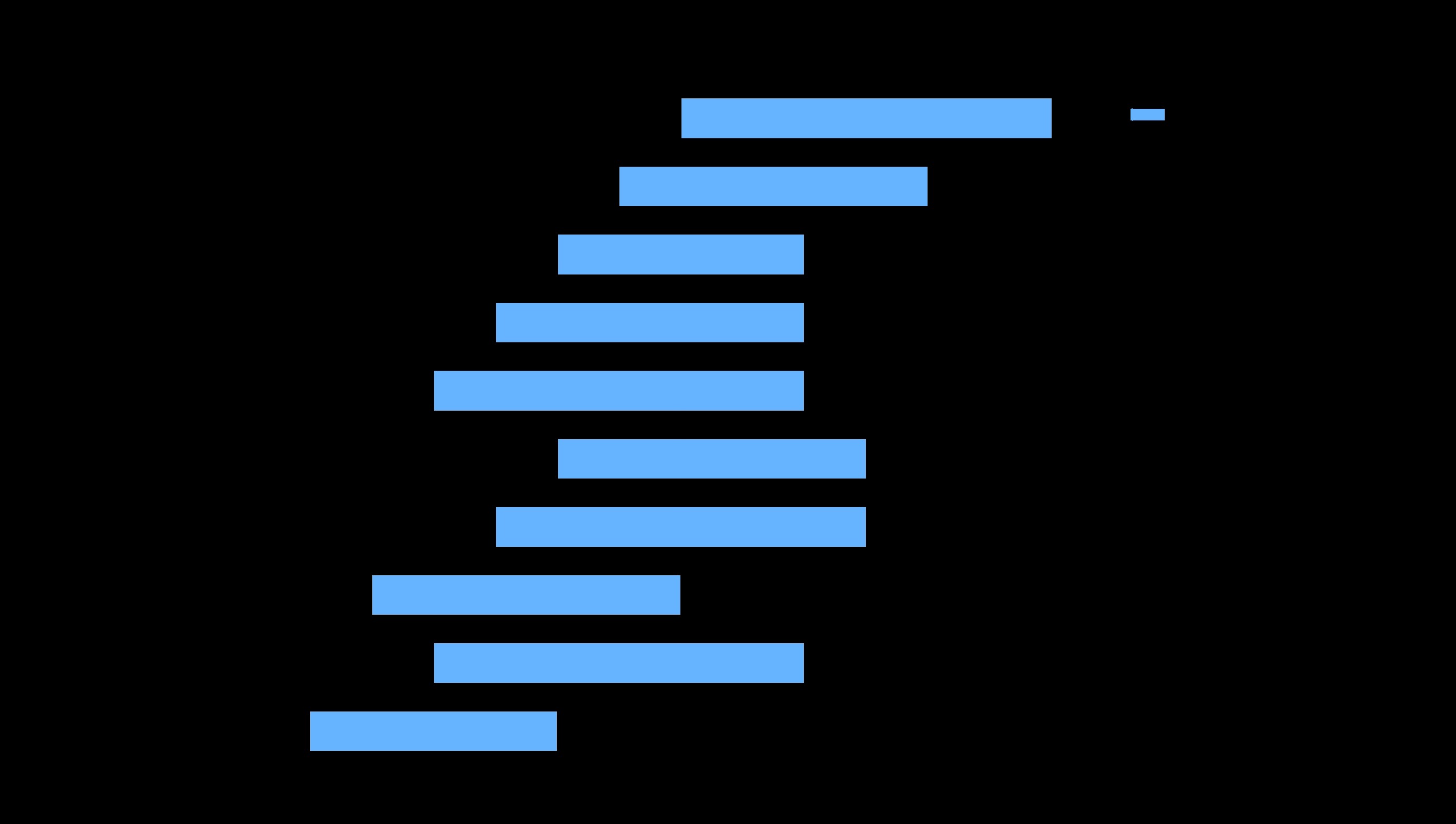} &
\includegraphics[width=\linewidth]{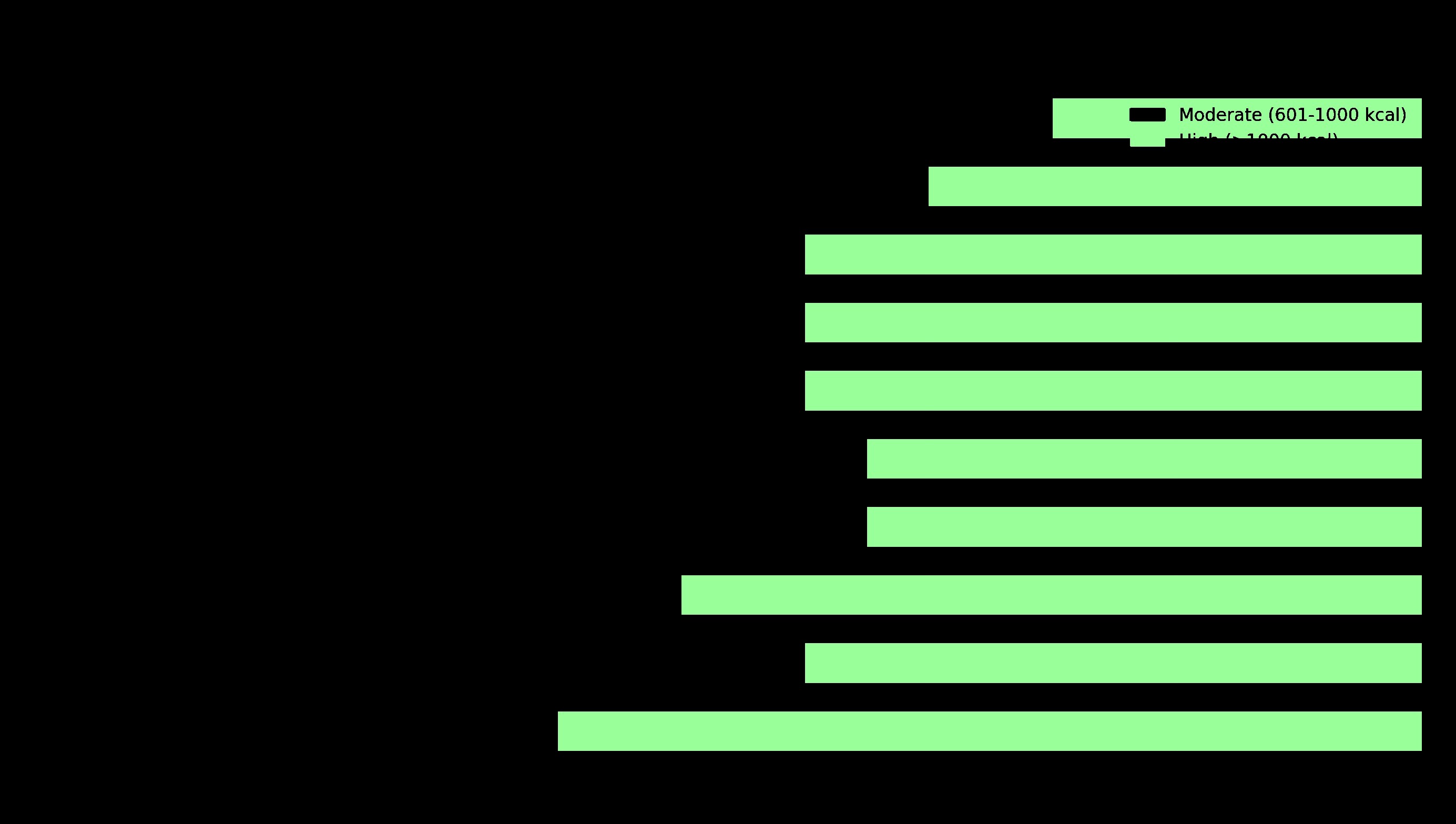} \\[-1pt]
{\scriptsize (b) \texttt{in\_range\_color}: isolate the low-calorie (pink) segment of every bar.} &
{\scriptsize (c) \texttt{in\_range\_color}: recover the moderate-calorie (blue) widths independently.} &
{\scriptsize (d) \texttt{in\_range\_color}: isolate the high-calorie (green) widths used in the numerator.}
\end{tabularx}

\vspace{1.5mm}

\begin{tcolorbox}[ovttrace]
\centering
\ovttool{in\_range\_color (Low)}
$\;\rightarrow\;$
\ovttool{in\_range\_color (Moderate)}
$\;\rightarrow\;$
\ovttool{in\_range\_color (High)}
$\;\rightarrow\;$
\ovttool{write\_file}
$\;\rightarrow\;$
\ovttool{exec}

\smallskip
\raggedright
The three masks return aligned bounding boxes for all ten categories.  For each category $c$, the model computes a ratio from the measured widths and then averages the ten ratios:
\[
    \frac{1}{10}\sum_{c=1}^{10}
    \frac{w_c^{\mathrm{High}}}
        {w_c^{\mathrm{Low}}+w_c^{\mathrm{Moderate}}}
    \;=\; \mathbf{1.062}.
\]
\end{tcolorbox}

\caption{A retained Chart trajectory.  Three clean color-range calls separate the low-, moderate-, and high-calorie segments of every stacked bar.  Their returned bounding boxes provide the measurements used in the final script.  This evidence raises the probe's avg@4 success from $0.50$ to $0.75$.}
\label{fig:case-chart-retained}
\end{figure*}

% Provenance: VinciCoder id 50327; session_50327.
% p0=0, ptool=1, correctness=html_vlm_score:95.00.
\begin{figure*}[p]
\centering
\small

\begin{tcolorbox}[
    ovtcase,
    colback=green!3,
    colframe=green!45!black
]
\textbf{\textsc{Retained} $\boldsymbol{\cdot}$ Web-to-HTML: render, diagnose, and revise}
\hfill
{\footnotesize $\bar p_0=0.00 \;\longrightarrow\; \bar p_{\mathrm{tool}}=1.00$, \textbf{$g=+1.00$}}

\smallskip
\textbf{Task.} Reproduce the reference webpage as a self-contained HTML/CSS file, using the prescribed placeholder for images.
\end{tcolorbox}

\begin{tabularx}{\textwidth}{@{}*{2}{>{\centering\arraybackslash}X}@{}}
\includegraphics[width=\linewidth]{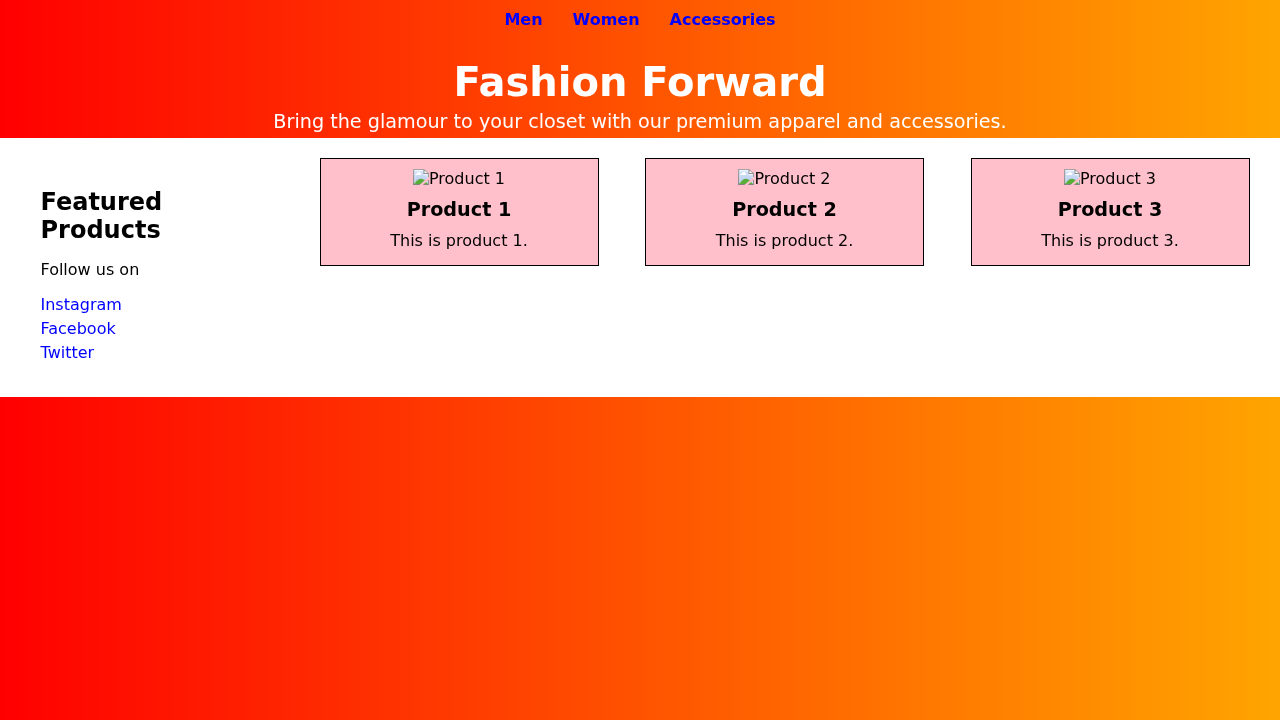} &
\includegraphics[width=\linewidth]{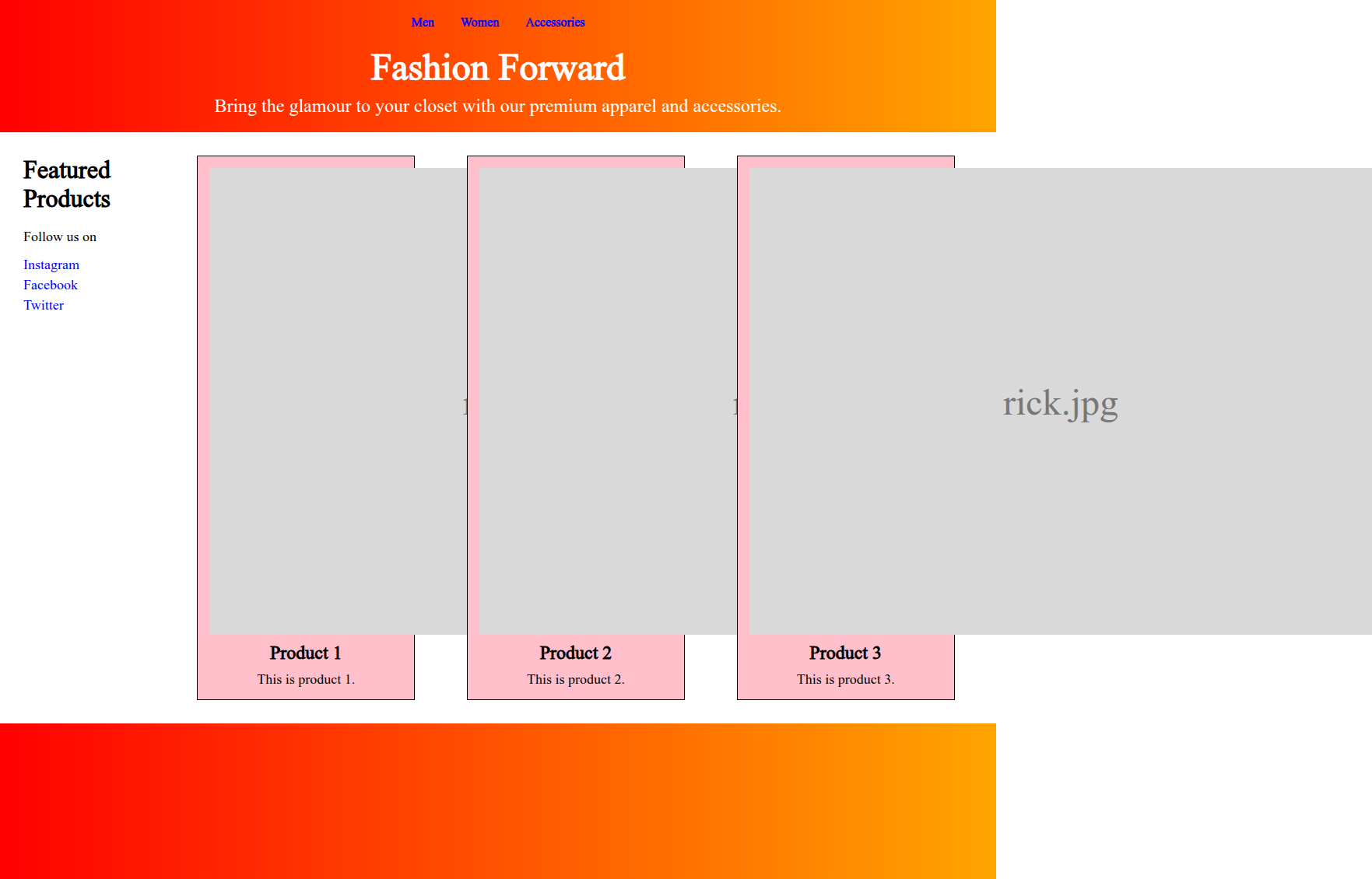} \\[-1pt]
{\scriptsize (a) Reference screenshot.} &
{\scriptsize (b) First \texttt{render\_html}: unconstrained placeholders overflow the product cards and push the page beyond the viewport.}
\end{tabularx}

\vspace{1.5mm}

\begin{tabularx}{\textwidth}{@{}*{2}{>{\centering\arraybackslash}X}@{}}
\includegraphics[width=\linewidth]{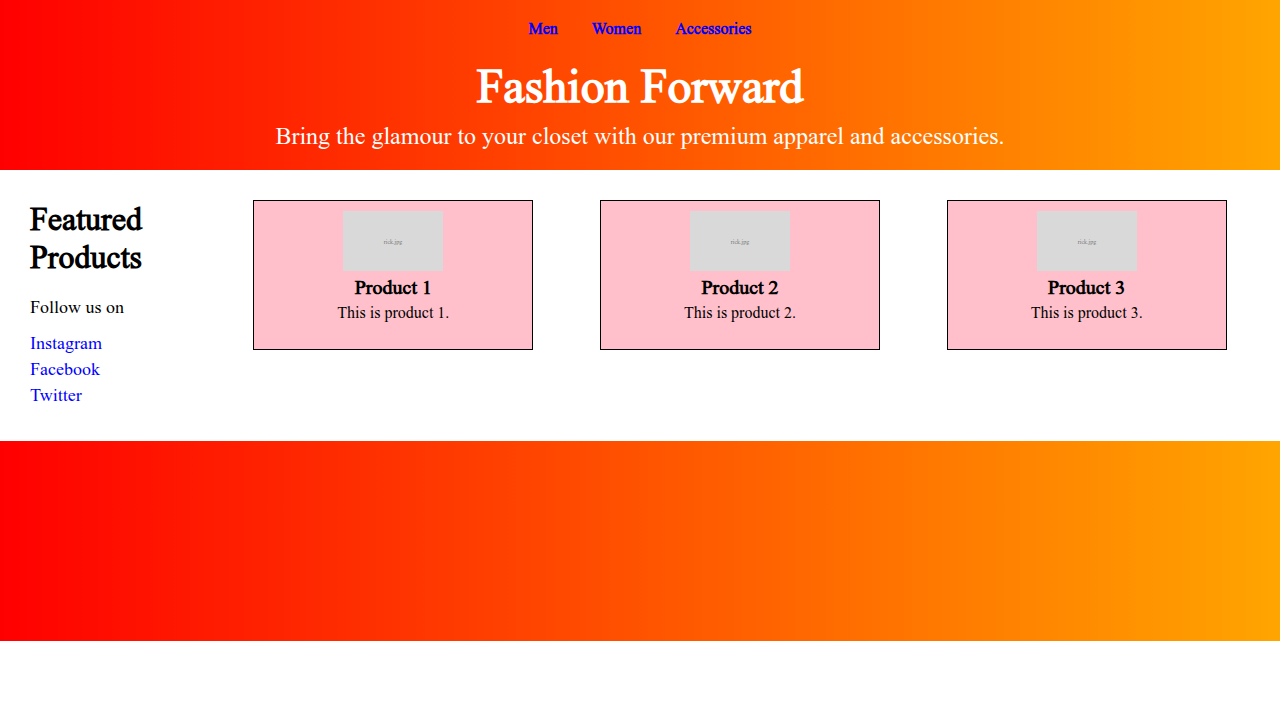} &
\includegraphics[width=\linewidth]{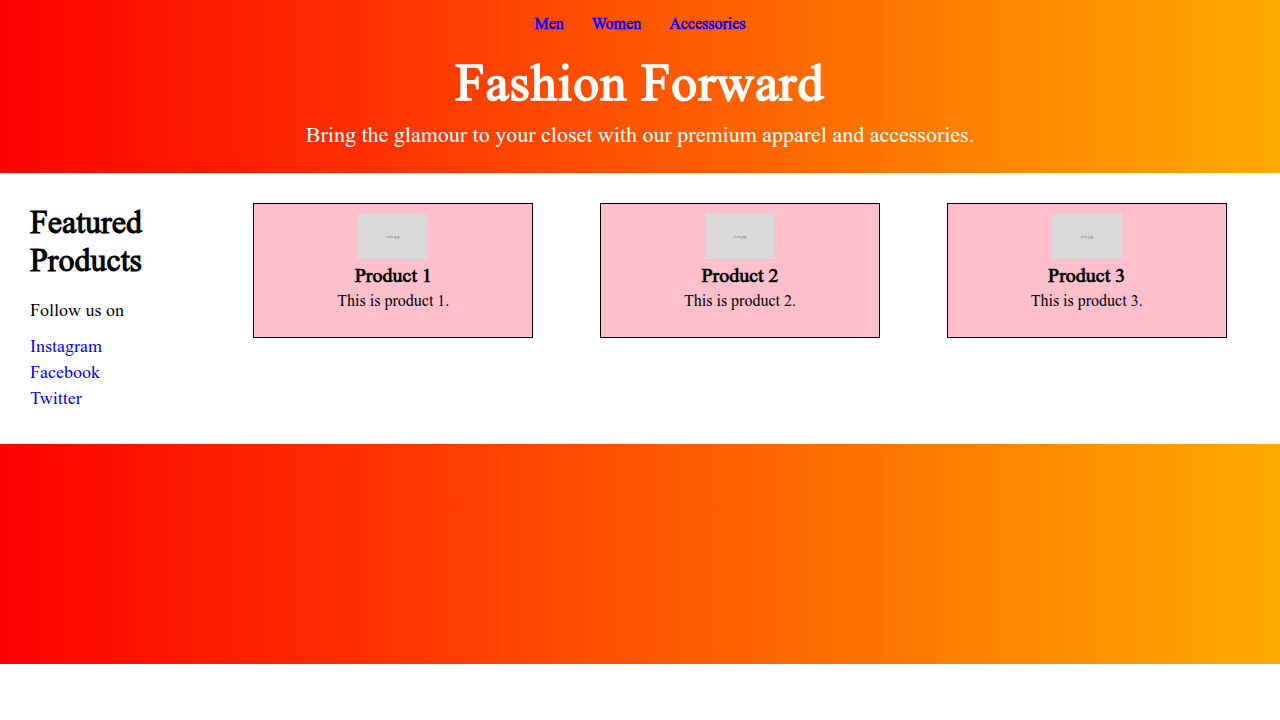} \\[-1pt]
{\scriptsize (c) First revision: fixed card and image dimensions restore the intended three-card layout.} &
{\scriptsize (d) Final render after spacing, typography, and footer refinements.}
\end{tabularx}

\vspace{1.5mm}

\begin{tcolorbox}[ovttrace]
\centering
\ovttool{write\_file}
$\;\rightarrow\;$
\ovttool{render\_html}
$\;\rightarrow\;$
\fcolorbox{red!55!black}{red!3}{%
    \parbox{0.27\linewidth}{\centering\scriptsize ``The product cards are too tall; the images are taking up too much space.''}}
$\;\rightarrow\;$
\ovttool{edit\_file}
$\;\rightarrow\;$
\ovttool{render\_html}

\smallskip
\raggedright
The rendered observation exposes a layout failure that is not visible in the HTML source alone.  The teacher responds to that specific failure, then re-renders to verify the correction; the final page receives an HTML-VLM outcome score of $95/100$.
\end{tcolorbox}

\caption{A retained Web-to-HTML trajectory, condensed to the decision-changing render--revise loop.  The first render reveals severe image overflow; the subsequent edit constrains the placeholders and recovers the reference's compact horizontal layout.  Later render--edit iterations refine spacing and typography.}
\label{fig:case-web-retained}
\end{figure*}

% Provenance: ChartVerse id 55059; session_55059.
% Outcome-correct (bold), but p0=.50, ptool=.25, hence excluded from OpenVisTool.
\begin{figure*}[p]
\centering
\small

\begin{tcolorbox}[
    ovtcase,
    colback=red!3,
    colframe=red!55!black
]
\textbf{\textsc{Rejected} $\boldsymbol{\cdot}$ Outcome-valid but no causal utility}
\hfill
{\footnotesize $\bar p_0=0.50 \;\longrightarrow\; \bar p_{\mathrm{tool}}=0.25$, \textbf{\textcolor{red!65!black}{$g=-0.25$}}}

\smallskip
\textbf{Question.} In the demographic group with the highest tertiary rate, what is the combined percentage with primary or secondary education?
\end{tcolorbox}

\begin{tabularx}{\textwidth}{@{}*{2}{>{\centering\arraybackslash}X}@{}}
\includegraphics[width=\linewidth]{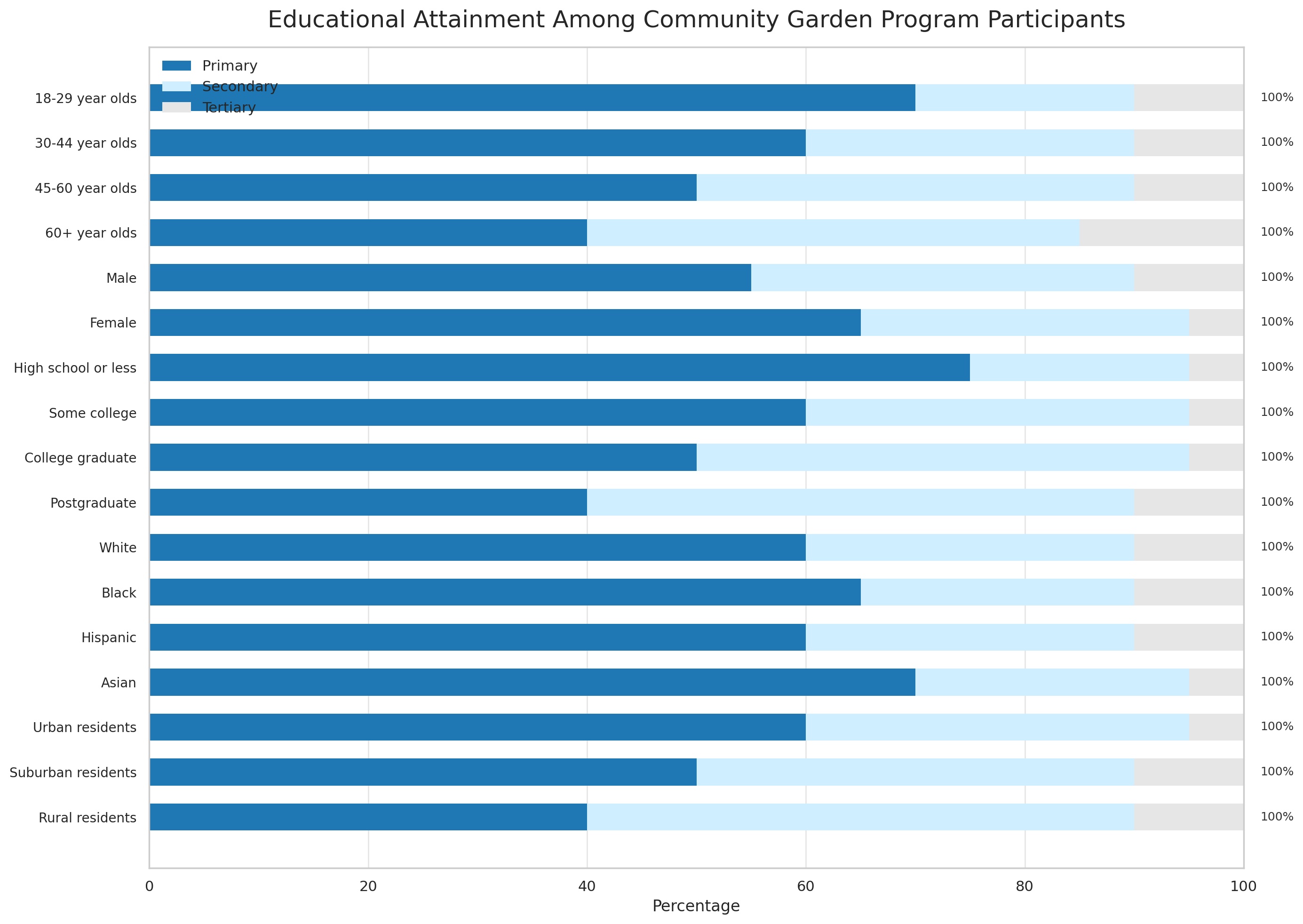} &
\includegraphics[width=\linewidth]{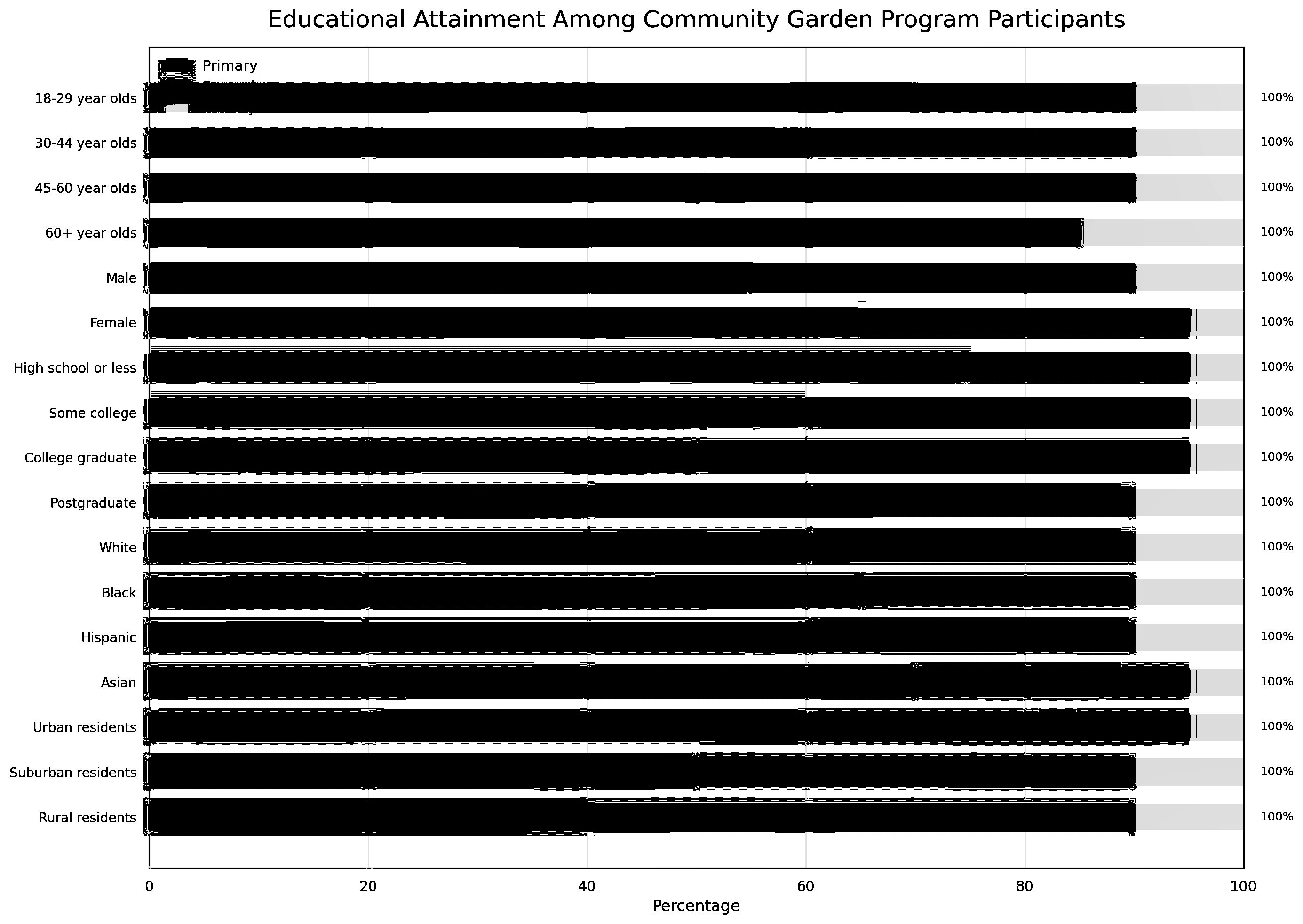} \\[-1pt]
{\scriptsize (a) Input.  The teacher already estimates the target group and $85\%$ from the raw/enhanced chart before receiving a useful localized observation.} &
{\scriptsize (b) Failed \texttt{in\_range\_color}: an over-broad low-saturation range selects essentially the entire chart (bbox $[0,0,1000,1000]$).}
\end{tabularx}

\vspace{1.5mm}

\begin{tabularx}{\textwidth}{@{}>{\centering\arraybackslash}X@{\hspace{4mm}}>{\centering\arraybackslash}X@{}}
\includegraphics[width=\linewidth]{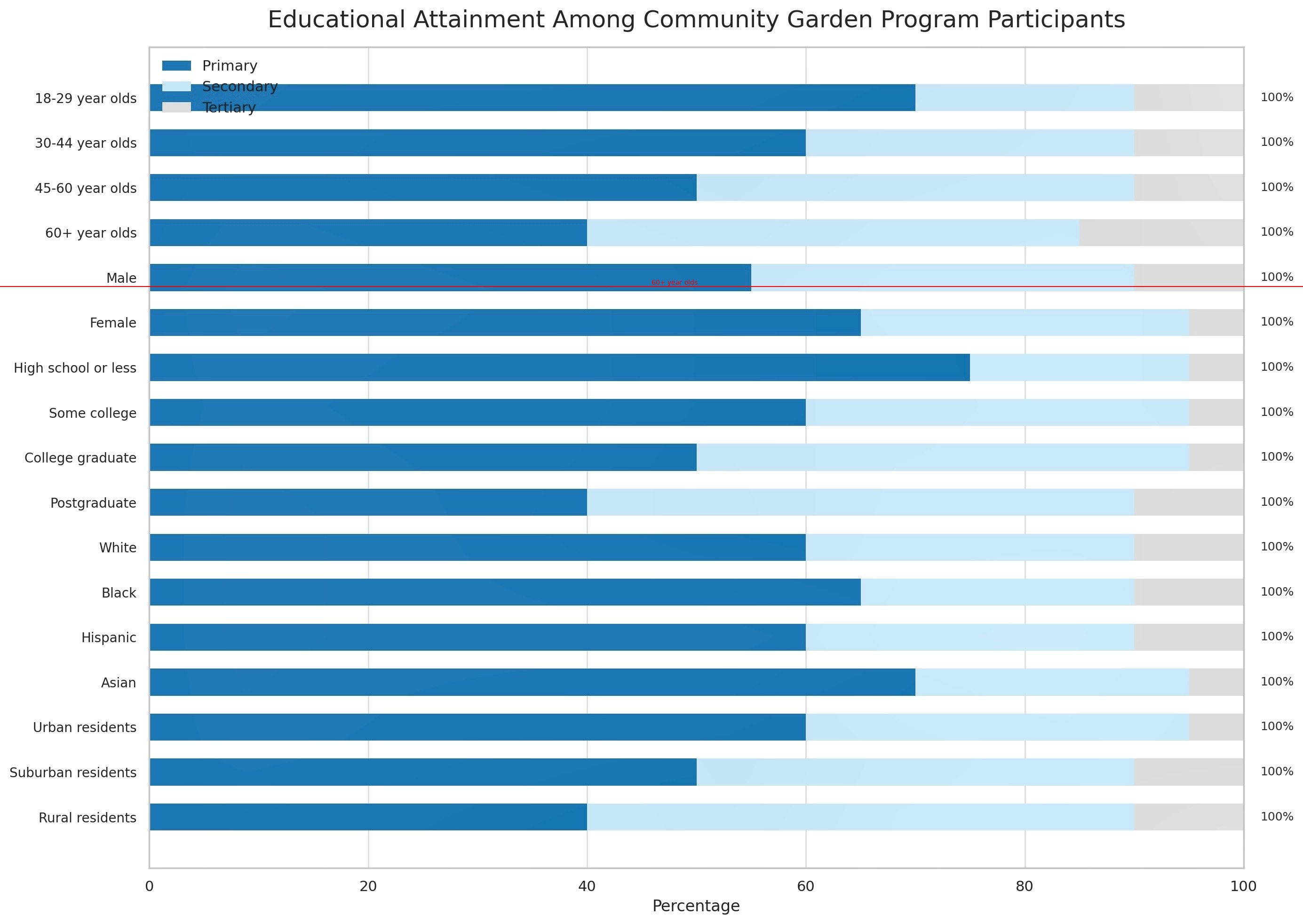} &
    \begin{tcolorbox}[
    ovttrace,
    colback=white,
    colframe=black!20,
    left=1mm,
    right=1mm,
    top=1mm,
    bottom=1mm
    ]
    \centering
    \includegraphics[width=\linewidth]{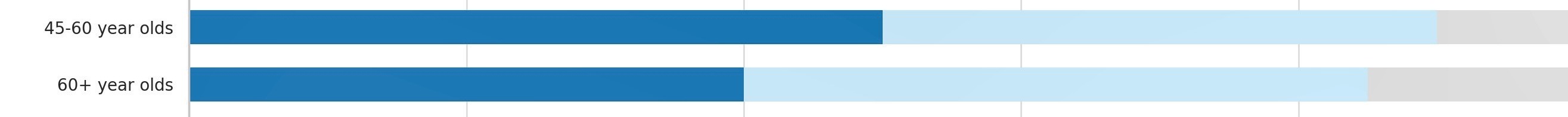}\\[-1pt]
    {\scriptsize (d) The late crop finally localizes the two relevant age rows.}

    \tcblower

    \raggedright\scriptsize
    \textbf{Late readout.} For the 60+ group, the crop gives $40\%_{\mathrm{primary}}+45\%_{\mathrm{secondary}}=85\%$. Crucially, the teacher had already inferred $85\%$ before this observation became available.
    \end{tcolorbox}
\\[-1pt]
{\scriptsize (c) Misplaced \texttt{draw\_line}: the guide labeled ``60+ year olds'' lands on the Male row.} & {}
\end{tabularx}

\vspace{1.5mm}

\begin{tcolorbox}[
    ovttrace,
    colback=red!2,
    colframe=red!45!black,
    colbacktitle=red!8,
    coltitle=black,
    fonttitle=\bfseries\scriptsize,
    title={Why causal utility filters this trajectory}
]
\begin{tabularx}{\linewidth}{@{}>{\centering\arraybackslash}X@{\hspace{3mm}}>{\centering\arraybackslash}p{0.31\linewidth}@{}}
    \ovttool{whole-chart mask}
    $\;\rightarrow\;$
    \ovttool{misplaced guide}
    $\;\rightarrow\;$
    \ovttool{late crop}
&
    \scriptsize
    \textbf{Outcome check:} \textsc{pass} ($85\%$)\\
    \textbf{Causal utility:} \textsc{reject} ($g=-0.25$)
\end{tabularx}

\smallskip
\scriptsize
The answer passes correctness, but the trace is dominated by uninformative or mislocalized observations; its only useful crop arrives after the answer has effectively been inferred.  Replaying the full trace lowers avg@4 from $0.50$ to $0.25$, so the sample is excluded from OpenVisTool.
\end{tcolorbox}

\caption{An outcome-correct trajectory removed by causal-utility filtering. The tool calls look superficially relevant, but the actual observations are uninformative or mislocalized and provide no counterfactual benefit.  This case illustrates why correctness-only filtering would retain spurious process supervision: the answer is correct, yet the tool trace decreases rather than improves the probe's success rate.}
\label{fig:case-rejected}
\end{figure*}

%% file: tables/tab_tool_complete.tex
\begin{table*}[p]
\centering
\small
\begin{tabular}{@{}p{0.22\textwidth}p{0.72\textwidth}@{}}
\toprule
\textbf{Tool} & \textbf{Description} \\
\midrule
\multicolumn{2}{@{}l}{\textit{General-purpose tools}} \\
\midrule
\texttt{read\_file} & Reads the contents of a file; image files are handled separately and returned as base64-encoded content. \\
\texttt{write\_file} & Writes content to a specified file. \\
\texttt{edit\_file} & Applies localized edits to an existing file. \\
\texttt{list\_dir} & Lists the contents of a specified directory. \\
\texttt{exec} & Executes a shell command. \\
\midrule
\multicolumn{2}{@{}l}{\textit{Geometric transformations}} \\
\midrule
\texttt{crop} & Crops a specified rectangular region from the image. \\
\texttt{rotate} & Rotates the image counter-clockwise by a specified angle, automatically expanding the canvas to fill the background. \\
\texttt{flip} & Flips the image horizontally or vertically. \\
\midrule
\multicolumn{2}{@{}l}{\textit{Annotation drawing}} \\
\midrule
\texttt{draw\_bbox} & Draws one or more rectangular bounding boxes, optionally with labels. \\
\texttt{draw\_circle} & Draws one or more circles, used to mark circular targets or point locations. \\
\texttt{draw\_line} & Draws one or more line segments, used to annotate axes, dividing lines, or reference lines. \\
\midrule
\multicolumn{2}{@{}l}{\textit{Feature extraction}} \\
\midrule
\texttt{in\_range\_color} & Generates a mask according to an HSV or BGR color range and returns information about the matching regions. \\
\texttt{connected\_components} & Extracts foreground connected components from a thresholded image, returning bounding boxes, centroids, and areas. \\
\texttt{detect\_edges} & Generates an edge map using the Sobel operator, highlighting object contours, text boundaries, and structural lines. \\
\texttt{find\_contours} & Extracts contours from a thresholded image, returning shape information such as area, perimeter, and bounding box. \\
\texttt{hough\_circles} & Detects circular targets via the Hough transform, suitable for bubble charts, radio buttons, or circular markers. \\
\texttt{hough\_lines} & Detects line segments via the probabilistic Hough transform, suitable for table lines, coordinate axes, and interface dividers. \\
\texttt{template\_match} & Locates positions in the image that match a template image or a cropped template region. \\
\midrule
\multicolumn{2}{@{}l}{\textit{Image processing}} \\
\midrule
\texttt{adjust\_brightness} & Adjusts image brightness and contrast, used to lighten, darken, or enhance overall visibility. \\
\texttt{enhance\_contrast} & Enhances contrast through histogram equalization, improving readability in low-contrast regions. \\
\texttt{grayscale} & Converts the image to grayscale, or further binarizes it to separate foreground from background. \\
\midrule
\multicolumn{2}{@{}l}{\textit{Visual verification}} \\
\midrule
\texttt{render\_html} & Renders an HTML file to a full-page screenshot using headless Chromium, enabling visual verification of generated HTML against a reference design. \\
\bottomrule
\end{tabular}
\caption{Complete toolset. General-purpose tools are shared across all task domains, while visual tools are grouped by functionality.}
\label{tab:tools-complete}
\end{table*}

%% file: tables/tab_domain_prompts.tex
% The wording below follows the domain-specific prompts used to construct
% OpenVisTool. Only Markdown-to-LaTeX typography has been changed.
\providecommand{\ovtpromptheading}[1]{%
  \par\smallskip\noindent\textbf{#1}\par\smallskip}
\providecommand{\ovtpromptitem}[1]{%
  \par\noindent\hangindent=1.25em\hangafter=1\textbullet\hspace{0.55em}#1\par}

\begin{table*}[p]
\centering
\footnotesize
\begin{tabularx}{\textwidth}{@{}>{\raggedright\arraybackslash}X@{}}
\toprule
\textbf{Chart VQA Tool-Use Instructions} \\
\midrule
{%
Below are the vision tools that frequently help on chart VQA. For each tool, a concrete trigger is listed---when the situation matches, call the corresponding tool instead of guessing from the raw image.

\ovtpromptheading{Geometric transforms}
\ovtpromptitem{\texttt{crop}: when the chart contains multiple subplots, inset views, dense legends, or small tick labels, or when only a specific region (a single subplot, a legend box, an axis area) is relevant. Zooming in reduces distraction and makes labels/values legible.}

\ovtpromptheading{Annotation / alignment aids}
\ovtpromptitem{\texttt{draw\_line}: when reading a value off an axis---place a vertical guide at the queried x-value or a horizontal guide at the queried y-value to avoid mis-aligning bar tops / line points with the axis ticks.}
\ovtpromptitem{\texttt{draw\_bbox}: when a specific region (a bar group, a legend entry, a highlighted area) must be tracked while cross-referencing it with the axis or legend.}
\ovtpromptitem{\texttt{draw\_circle}: when pointing to a single data point (a scatter marker, a line peak, a pie slice) to confirm it is the one the question asks about.}

\ovtpromptheading{Contrast / readability enhancement}
\ovtpromptitem{\texttt{enhance\_contrast}: when grid lines, low-contrast bars/lines, small tick labels, or compressed chart details are hard to read.}

\ovtpromptheading{Color-based lookup}
\ovtpromptitem{\texttt{in\_range\_color}: when the question depends on identifying a category, legend color, series color, or colored bar/line/area, or when estimating how much of a chart region belongs to a specific color. Prefer HSV ranges for robust selection under anti-aliasing/compression. Pass \texttt{region} to restrict matching to the plot area so legends, titles, and surrounding decorations are excluded.}

\ovtpromptheading{Compute}
\ovtpromptitem{If the question involves calculation (sum, mean, ratio, ranking, percentage change, slope, etc.), use \texttt{exec} or \texttt{write\_file} to create and run a Python script instead of relying on mental math.}
} \\
\bottomrule
\end{tabularx}
\caption{Domain-specific teacher-rollout prompt for Chart.}
\label{tab:prompt-chart}
\end{table*}

\begin{table*}[p]
\centering
\footnotesize
\begin{tabularx}{\textwidth}{@{}>{\raggedright\arraybackslash}X@{}}
\toprule
\textbf{Table VQA Tool-Use Instructions} \\
\midrule
{%
Below are the vision tools that frequently help on table VQA. For each tool, a concrete trigger is listed---when the situation matches, call the corresponding tool instead of guessing from the raw image.

\ovtpromptheading{Geometric transforms}
\ovtpromptitem{\texttt{crop}: when only a specific cell, row block, column block, header region, or small text is relevant, or when the image also contains surrounding captions, footnotes, or other tables. Zooming in reduces distraction and lets you read fine digits/units reliably.}

\ovtpromptheading{Annotation / alignment aids}
\ovtpromptitem{\texttt{draw\_bbox}: when you need to highlight and track a specific target cell or a set of candidate cells while cross-checking row label $\times$ column header $\times$ value.}
\ovtpromptitem{\texttt{draw\_line}: when you must align a row with a column across a wide table; drawing a horizontal line across the target row or a vertical line down the target column avoids off-by-one row/column mismatches.}

\ovtpromptheading{Contrast / readability enhancement}
\ovtpromptitem{\texttt{enhance\_contrast}: when the table has low contrast (faded scans, light-gray zebra stripes, watermark bleed-through), small digits are hard to read, or cell backgrounds differ in subtle shades that interfere with text.}

\ovtpromptheading{Color-based lookup}
\ovtpromptitem{\texttt{in\_range\_color}: when the question depends on cells of a specific color (highlighted rows, conditional formatting, colored status cells); the HSV range mask isolates them and returns per-component bboxes.}

\ovtpromptheading{Compute}
\ovtpromptitem{If the question involves calculation (sum, mean, ratio, ranking, percentage change, etc.), use \texttt{exec} or \texttt{write\_file} to create and run a Python script instead of relying on mental math.}
} \\
\bottomrule
\end{tabularx}
\caption{Domain-specific teacher-rollout prompt for Table.}
\label{tab:prompt-table}
\end{table*}

\begin{table*}[p]
\centering
\footnotesize
\begin{tabularx}{\textwidth}{@{}>{\raggedright\arraybackslash}X@{}}
\toprule
\textbf{Visual Search Tool-Use Instructions} \\
\midrule
{%
Below are the vision tools that deliver the largest gains on visual search / grounding / counting / attribute-verification questions. When the situation matches a trigger, call the tool instead of guessing from the raw image.

\ovtpromptitem{\texttt{crop}: when the target is a small object in a high-resolution image, partially occluded, distant, or surrounded by clutter. Zoom into the candidate region to verify fine attributes (color, shape, text, fine-grained category) before answering.}
\ovtpromptitem{\texttt{draw\_bbox}: when the question depends on locating one or more candidate objects, verifying a spatial relation (``is A to the left of B?''), or keeping track of multiple candidates during search. Drawing bboxes helps avoid missed or duplicate counting in crowded scenes.}
\ovtpromptitem{\texttt{in\_range\_color}: when the target is defined primarily by color (``the red car'', ``the blue backpack'', ``all yellow flowers''); the HSV mask isolates matching pixels and returns per-component bboxes that you can then count or verify.}
\ovtpromptitem{\texttt{enhance\_contrast}: when the image is low-contrast (foggy, hazy, overcast, low-light indoor) and candidate objects blend into the background; CLAHE on LAB often reveals hidden targets without shifting colors.}
\ovtpromptitem{\texttt{adjust\_brightness}: when the image is clearly too dark (night scenes, shadows) or too bright (overexposed sky, white backgrounds with blown highlights) and the target is lost in the extreme; tune \texttt{alpha}/\texttt{beta} to recover details.}
\ovtpromptitem{\texttt{exec}: when counting or arithmetic over detected items is required (e.g. ``how many more red cars than blue cars''), collate the per-detection JSON payloads and compute the answer rather than counting by eye.}
} \\
\bottomrule
\end{tabularx}
\caption{Domain-specific teacher-rollout prompt for Visual Search.}
\label{tab:prompt-visual-search}
\end{table*}

\begin{table*}[p]
\centering
\footnotesize
\begin{tabularx}{\textwidth}{@{}>{\raggedright\arraybackslash}X@{}}
\toprule
\textbf{GUI Grounding Tool-Use Instructions} \\
\midrule
{%
The input is a screenshot of a GUI, and the query asks you to locate a specific UI element (e.g. ``click the Submit button'', ``find the search bar'', ``where is the settings icon?''). Your job is to locate that element precisely and return its click position as the final answer.

\ovtpromptheading{Required workflow}
\textbf{First, use visual tools to find and verify the target.} Do not guess the coordinate from the raw screenshot. Always confirm the element's position by at least one of the tools below before committing to a final click. The coordinate you finally emit must be the \textbf{center} of the target element, not its edge or corner. Each tool call should have a clear hypothesis to confirm or reject---only call when it actually reduces ambiguity.

\smallskip
\textbf{Finally, produce the answer as a single \texttt{computer\_use} tool\_call with \texttt{action: "*\_click"}.} The \texttt{coordinate: [x, y]} must point at the center of the target element. The screen is treated as a 1000$\times$1000 canvas, so coordinates are in the normalized \texttt{[0, 1000]} space---never return raw pixel coordinates from the original image. Emit exactly one \texttt{computer\_use} call; it terminates the trajectory and is treated as your final output. Do not follow it with any other tool call or free-form text.

\ovtpromptheading{Pre-answer visual tools}
\ovtpromptitem{\texttt{crop}: zoom into the candidate region to read small labels, verify icons, or disambiguate between nearby elements. Especially important when the target is a small icon, a list item, a toolbar button, or text inside a dense layout.}
\ovtpromptitem{\texttt{draw\_bbox}: when multiple candidate elements exist (``the third item in the list'', ``the button next to X''), or when you want to visually confirm in advance that the target you plan to click is the intended element.}
\ovtpromptitem{\texttt{in\_range\_color}: when the target is primarily identified by color (``the red alert'', ``the green confirm button'') and shape alone is ambiguous; the HSV mask returns per-component bboxes.}
\ovtpromptitem{\texttt{enhance\_contrast} / \texttt{adjust\_brightness}: when the screenshot is dim, washed out, or has heavy dark-mode shadows that hide the target.}
\ovtpromptitem{\texttt{detect\_edges} / \texttt{find\_contours}: when the target is defined by a thin outline (icon silhouette, table border, dividing line) that is hard to separate visually.}
} \\
\bottomrule
\end{tabularx}
\caption{Domain-specific teacher-rollout prompt for GUI Grounding.}
\label{tab:prompt-gui-grounding}
\end{table*}

\begin{table*}[p]
\centering
% \scriptsize
\footnotesize
\begin{tabularx}{\textwidth}{@{}>{\raggedright\arraybackslash}X@{}}
\toprule
\textbf{HTML Code Generation Tool-Use Instructions} \\
\midrule
{%
The task is to reproduce the webpage in the reference screenshot as faithfully as possible by emitting a single self-contained HTML document. \textbf{Use the tools below to verify and correct your draft instead of relying on a one-shot guess}---a draft that ``looks right'' in your head almost always diverges from the reference once rendered.

\ovtpromptheading{Required workflow}
The trajectory must follow this shape; do not collapse steps:

\begin{enumerate}[leftmargin=1.5em,itemsep=1.5pt,topsep=2pt,parsep=0pt]
\item \textbf{(Optional) Inspect the reference with vision tools first.} Call these \emph{before} writing any HTML when they actually reduce ambiguity:
  \begin{itemize}[leftmargin=1.5em,itemsep=0pt,topsep=1pt,parsep=0pt]
  \item \texttt{crop}---when the screenshot is tall, dense, or has small text you can't read at thumbnail level. Crop a single region (header / hero / nav / cards / footer) and look at it in isolation.
  \item \texttt{in\_range\_color} / \texttt{sample\_color}---when you would otherwise guess a hex value for a \texttt{background-color}, brand accent, button fill, or border. Sample the actual pixels and lock the palette before writing CSS.
  \item \texttt{enhance\_contrast} / \texttt{detect\_edges}---only when the layout edges or borders are genuinely hard to see; skip otherwise.
  \end{itemize}
  Skip this step on visually simple pages---but explain in your thinking why you can skip it. Don't call these tools just to seem thorough.

\item \textbf{Write the first draft to a file with \texttt{write\_file}.} Inline all CSS in a \texttt{<style>} block. Keep \texttt{rick.jpg} placeholders literally as-is. Use a clear filename (e.g. \texttt{index.html}).

\item \textbf{Call \texttt{render\_html} on the file you just wrote.} Pass its \texttt{path}---\texttt{render\_html} reads the HTML from disk, so you must \texttt{write\_file} before the first \texttt{render\_html} call. Whenever you can read the reference's pixel size off the original, pass it as \texttt{viewport\_width} / \texttt{viewport\_height} so layout breakpoints and full-page heights match. The tool returns the rendered screenshot---visually compare it to the reference end-to-end.

\item \textbf{Name the discrepancies concretely.} After every \texttt{render\_html} call, in your thinking, list the specific diffs you can see---e.g. ``nav links not horizontal'', ``card padding too small'', ``hero image is left-aligned but should be centered'', ``primary button is too saturated''. If you cannot name any concrete diff, the draft is good enough---go to step 6.

\item \textbf{Patch the HTML with \texttt{edit\_file}, then re-render the same file.} Prefer \texttt{edit\_file} over rewriting the whole file with \texttt{write\_file}---large rewrites destroy the parts that were already correct and waste tokens. After each patch, call \texttt{render\_html} on the same \texttt{path} again (the tool always reads the latest contents from disk) and re-evaluate. Iterate steps 4--5 until either the rendered screenshot is visually consistent with the reference, or a further patch is no longer closing the gap.

\item \textbf{Submit the final HTML in the assistant message.} Quote the full HTML once, then stop.
\end{enumerate}

\ovtpromptheading{Pitfalls to avoid}
\ovtpromptitem{\textbf{At least one \texttt{render\_html} call is required} for every trajectory---even on simple pages. The closed loop is the whole point.}
\ovtpromptitem{\textbf{Patch with \texttt{edit\_file}, don't rewrite via \texttt{write\_file}.} Rewriting the whole HTML between iterations destroys the parts that were already correct and wastes tokens; \texttt{write\_file} is only for the initial draft.}
\ovtpromptitem{\textbf{Do not reference external assets} (CDN images, Google Fonts, remote stylesheets)---the sandbox can't reach them, the render will show broken images, and the next comparison will be misleading. Inline styles, keep \texttt{rick.jpg}-style placeholders verbatim, and rely on web-safe font stacks.}
} \\
\bottomrule
\end{tabularx}
\caption{Domain-specific teacher-rollout prompt for Web-to-HTML.}
\label{tab:prompt-web-to-html}
\end{table*}